\documentclass{article}

\PassOptionsToPackage{numbers, compress, sort}{natbib}

\usepackage[preprint]{neurips_2023}

\usepackage[utf8]{inputenc} 
\usepackage[T1]{fontenc}    
\usepackage{url}            
\usepackage{booktabs}       
\usepackage{amsfonts}       
\usepackage{nicefrac}       
\usepackage{microtype}      
\usepackage{xcolor}         
\usepackage{amsmath}
\usepackage{graphicx}
\usepackage{makecell}
\usepackage{subfig}

\usepackage[hidelinks]{hyperref}       
\usepackage{multirow}
\usepackage{bbm}
\usepackage{makecell}
\usepackage{adjustbox}
\usepackage{float}
\usepackage{xr}
\usepackage{bm}
\usepackage{epigraph}

\usepackage{caption}
\usepackage{appendix}
\AtBeginEnvironment{subappendices}{%
    \chapter*{Appendices}
    \addcontentsline{toc}{chapter}{Appendices}
    \counterwithin{figure}{section}
    \counterwithin{table}{section}
}

\AtEndEnvironment{subappendices}{%
    \counterwithin{figure}{chapter}
    \counterwithin{table}{chapter}
}

\title{Integrating Random Effects in Variational Autoencoders for Dimensionality Reduction of Correlated Data}

\author{%
  Giora Simchoni \\
  Department of Statistics\\
  Tel Aviv University\\
  Tel Aviv, Israel, 69978 \\
  \texttt{gsimchoni@tauex.tau.ac.il} \\
   \And
   Saharon Rosset \\
   Department of Statistics\\
   Tel Aviv University\\
   Tel Aviv, Israel, 69978 \\
   \texttt{saharon@tauex.tau.ac.il} \\
}

\begin{document}

\maketitle

\begin{abstract}
  Variational Autoencoders (VAE) are widely used for dimensionality reduction of large-scale tabular and image datasets, under the assumption of independence between data observations. In practice, however, datasets are often correlated, with typical sources of correlation including spatial, temporal and clustering structures. Inspired by the literature on linear mixed models (LMM), we propose LMMVAE -- a novel model which separates the classic VAE latent model into fixed and random parts. While the fixed part assumes the latent variables are independent as usual, the random part consists of latent variables which are correlated between similar clusters in the data such as nearby locations or successive measurements. The classic VAE architecture and loss are modified accordingly. LMMVAE is shown to improve squared reconstruction error and negative likelihood loss significantly on unseen data, with simulated as well as real datasets from various applications and correlation scenarios. It also shows improvement in the performance of downstream tasks such as supervised classification on the learned representations.
\end{abstract}

\section{Introduction} \label{intro_lmmvae}
In recent years, variational autoencoders (VAE, \cite{VAE2013}) have become ubiquitous in machine learning research, owing to their scalability and tendency to find rich non-linear representations. Though originally demonstrated to work on image data, VAE have been used extensively for performing non-linear dimensionality reduction for large-scale tabular datasets in fields such as healthcare and bioinformatics \cite{Wei2021Review}, finance \cite{Bergeron125}, neuroimaging \cite{Ning2020} and education \cite{Curi2019}. Such datasets often exhibit a clear pattern of dependency between observations. To illustrate, consider the UK Biobank dataset \cite{UKB}, in which a cohort of almost half a million adult individuals across the UK are followed for health-related indicators, since 2006. Two blood pressure measurements could have temporal dependence (repeated measures in time from individual $i$), spatial dependence (individual $i$ resides in the same neighborhood as individual $j$), cluster dependence (individuals $i$ and $j$ frequent the same doctor, one of tens of thousands) or any combination of these and others.

Yet, it is assumed by design of the classical VAE that the latent variables (LV) underlying the data are \emph{independent} between observations, thus limiting VAE ability to reach optimal representations for such correlated data. Moreover, while many solutions for handling correlated data exist for VAE's linear ancestor principal component analysis (PCA, \citep[see e.g.][]{jolliffe2002}), there is relatively little previous work dedicated to modifying VAE to account for dependency structures in datasets. See Section~\ref{sec:background} for more details.

In this work we aim to enable VAE to incorporate structured correlation between observations by taking inspiration from probabilistic PCA (PPCA, \cite{PPCA}) and the field of linear mixed models (LMM, \citep[see e.g.][]{McCulloch2008}). LMM are traditionally used in regression and classification settings, where correlated data are ubiquitous. Inherent to LMM is the use of random effects (RE), which are random additions to the fixed effects (FE) of some of the regression model's features. These RE reflect correlation between observations from the same clusters, similar times, near locations etc. The model presented here modifies the PPCA formulation in two ways. First, the LV $u$ are no longer linearly related to the original high-dimensional features, but through a non-linear function $f$, suitable for fitting with VAE. The fitted function $f(u)$ represents the ``fixed'' part of the latent relation. Second, we add a ``random'' part much like in LMM, reflecting the similarity between correlated observations. We then modify the VAE architecture and loss to fit our model, hence its title LMMVAE. LMMVAE is demonstrated to perform consistently well on all of the aforementioned dependency types and combinations of them, on simulated as well as real datasets. We show superior performance of LMMVAE in comparison to other SOTA methods and to ignoring the dependency, as judged by squared reconstruction error on unseen data, negative log-likelihood (NLL) loss, downstream tasks on the latent representations and qualitative plots.

The rest of this paper is organized as follows: in Section~\ref{sec:background} we briefly review PPCA, VAE and previous attempts at extending them to handle correlated data. In Section~\ref{sec:LMMVAE} we describe our main contribution LMMVAE and show how to adapt it to various correlation structures. In Section~\ref{sec:experiments} we show excellent results when comparing LMMVAE to some of the methods from Section~\ref{sec:background} on simulated and real tabular and image datasets, before concluding in Section~\ref{sec:conclusion}.

\section{Background and Related Methods}
\label{sec:background}
\subsection{Probabilistic PCA}
Let $\mathbf{X}$ of order $n \times p$ represent a data matrix of $n$ observations measured over $p$ features. The PPCA model \cite{PPCA} assumes each of the $p$-dimensional observations (rows) $\mathbf{x}_i$ ($i=1,\dots, n$) is linearly generated from a set of $d$ independent LV $\mathbf{u}_i$ coming from a $\mathcal{N}(\mathbf{0}, \mathbf{I}_d)$ distribution, where typically $d \ll p$:
\begin{equation}
\label{eq:PPCA}
    \mathbf{x}_i = \mathbf{W}\mathbf{u}_i + \bm{\mu} + \bm{\varepsilon}_i
\end{equation}
Here $\mathbf{W}$ is a $p \times d$ loadings matrix, $\bm{\mu}$ is a vector of means, and $\bm{\varepsilon}_i$ is Gaussian noise from a $\mathcal{N}(\mathbf{0}, \bm{\Sigma})$ distribution, where $\bm{\Sigma}$ is a $p \times p$ covariance matrix. Hence, we can write the $\mathbf{x}_i$ marginal distribution as $\mathcal{N}(\bm{\mu}, \mathbf{W}\mathbf{W}^T + \bm{\Sigma})$. If $\bm{\Sigma} = \sigma^2\mathbf{I}_p$ then \cite{PPCA} have shown that as $\sigma^2 \to 0$ the maximum likelihood $\mathbf{W}$ loadings matrix is that derived by PCA up to rotation, by extracting the first $d$ eigenvectors of the $\mathbf{X}^T\mathbf{X}$ covariance matrix, assuming $\mathbf{X}$ is centered.

Many attempts have been made at generalizing PCA and PPCA to fit correlated data and non-linear relations between the latent and output spaces. Functional PCA \citep[see e.g.][]{ramsay2005functional} assumes the principal components are in fact \emph{curves}, or functions of time $t$, which might be more fitting for time series data. Kernel PCA \cite{KPCA} applies the kernel trick to PCA by replacing the covariance matrix $\mathbf{X}\mathbf{X}^T$ used by the dual form of PCA with a positive semi-definite $n \times n$ kernel matrix $\mathbf{K}$ before applying PCA as usual, thus allowing to compute PCA on arbitrarily high-dimensional non-linear mappings of $\mathbf{X}$. Finally, Gaussian process LV model (GP-LVM) combines PPCA and Kernel PCA by adopting the dual view of PPCA and applying the kernel trick on it \cite{Lawrence05}. 

\subsection{Variational Autoencoders}
The most popular non-linear scalable version of PCA in the deep learning era, however, has been autoencoders and their recent variational-inference successors, variational autoencoders \cite{hinton2006, VAE2013, VAE2019}. It is well-known that VAE could be seen as generalizing the linear relation in \eqref{eq:PPCA} with a general function $f: \mathbb{R}^d \to \mathbb{R}^p$ which we fit with a deep neural network (DNN):
\begin{equation}
\label{eq:VAE-as-PPCA}
    \mathbf{x}_i = f(\mathbf{u}_i) + \bm{\varepsilon}_i
\end{equation}
In keeping with the VAE Bayesian phrasing, we say $\mathbf{x}$ is generated from latent $\mathbf{u}$ by a distribution $p_\theta(\mathbf{x}|\mathbf{u})$, parameterized by some unknown parameters $\theta$. When this is multiplied by $\mathbf{u}$'s prior distribution $p_\theta(\mathbf{u})$ it gives the full deep latent variable model (DLVM): $p_\theta(\mathbf{x}, \mathbf{u}) = p_\theta(\mathbf{x}|\mathbf{u})p_\theta(\mathbf{u})$. Extracting the posterior of $\mathbf{u}$ given the data $p_\theta(\mathbf{u}|\mathbf{x})$ involves calculating the marginal likelihood $p_\theta(\mathbf{x}) = \int p_\theta(\mathbf{x}|\mathbf{u})p_\theta(\mathbf{u})\,d\mathbf{u}$ after integrating out the LV $\mathbf{u}$, but $p_\theta(\mathbf{x})$ is usually intractable. VAE approximates $p_\theta(\mathbf{u}|\mathbf{x})$ with an \emph{encoder} model $q_\phi(\mathbf{u}|\mathbf{x})$ which is typically of simpler form. By doing so, the log marginal likelihood can be approximately maximized by maximizing the evidence lower bound (ELBO):
\begin{equation}
\label{eq:ELBO1}
    \mathcal{L}_{\theta, \phi}(\mathbf{x}) = \mathbb{E}_{q_\phi(\mathbf{u}|\mathbf{x})}[\log p_\theta(\mathbf{x}|\mathbf{u})] - D_{KL}[q_\phi(\mathbf{u}|\mathbf{x}) || p_\theta(\mathbf{u})],
\end{equation}
 where the second term is the Kullback-Leibler (KL) divergence between the encoder $q_\phi(\mathbf{u}|\mathbf{x})$ and the prior $p_\theta(\mathbf{u})$, and often it is multiplied by a hyperparameter $\beta$ ($\beta$-VAE) which has been shown to result in orthogonal LV \cite{betaVAE, Rolinek2018}. In practice one often minimizes the negative ELBO (or: NLL) which can be simplified to a squared reconstruction term $\|\mathbf{x} - f(\mathbf{u})\|^2$ for Gaussian log-likelihood, and an additional term which can be interpreted as regularization \cite{VAE2019}.

There are a few extensions to VAE aimed at dealing with correlated data, some of which we compare our results to. Variational recurrent autoencoders (VRAE) have been shown to produce useful representations for musical data and short sentences \cite{VRAE, Bowman2015}. VRAE are essentially a change of architecture to VAE -- utilizing LSTM cells in the encoder/decoder networks instead of multi-layer perceptrons (MLP). A more theoretically justified approach is that of Gaussian process prior variational autoencoders (GPPVAE) and a more recent version using SGD, scalable variational GP VAE (SVGPVAE) \cite{GPPVAE, SVGPVAE}. The motivation behind these approaches is similar to ours: accounting for correlation between observations. However, while LMMVAE is inspired by LMM to separate the generating model into fixed and random parts, GPPVAE and SVGPVAE strive to put a GP prior on the LV $\mathbf{u}$, instead of the $\mathcal{N}(\mathbf{0}, \mathbf{I})$ factorized prior in regular VAE. Assume $Q$ unique entities are repeatedly measured for a total of $n$ observations, that is data $\mathbf{X}$. Furthermore, there is low-dimensional auxiliary data $\mathbf{Y}$ on these entities (e.g., the timestamp of a rotating object's frame). The LV $\mathbf{u}$ are generated from $\mathbf{Y}$ with a GP prior: $p(\mathbf{u}|\mathbf{y}) = \prod_{l=1}^d \mathcal{N}(\mathbf{0}, \mathbf{K}(\mathbf{y}, \mathbf{y}))$, then $\mathbf{X}$ is generated from $\mathbf{u}$ as in model ~\eqref{eq:VAE-as-PPCA}, the standard VAE. While GPPVAE has complexity of $\mathcal{O}(n)$ for a single gradient iteration and is therefore unsuitable for large datasets, SVGPVAE utilizes the theory of inducing points to reach complexity of $\mathcal{O}(b^2m + m^3)$, where $b$ is the mini-batch size, and $m$ is the number of inducing points. The SVGPVAE setting, however, may be well suited for longitudinal datasets, but not for scenarios where in a tabular dataset there are spatial features such as longitude/latitude, with no additional auxiliary data $Y$ on locations. Indeed, both GPPVAE and SVGPVAE have been demonstrated to work on artificial image data only, for example to predict the next image for a rotating MNIST digit, whereas the focus of this study is more on tabular datasets which express various types of correlation between observations. Below LMMVAE outperforms SVGPVAE where relevant. For completeness, we also mention hierarchical variational models (HVM) and auxiliary deep generative models (ADGM) \cite{HVM, ADGM}. These models allow for a more expressive variational distribution, either by adding a prior on the variational parameters, or by adding auxiliary LV $\mathbf{z}$ and writing the posterior distribution as $q_\phi(\mathbf{u},\mathbf{z}|\mathbf{x}) = q_\phi(\mathbf{z}|\mathbf{x})q_\phi(\mathbf{u}|\mathbf{z}, \mathbf{x})$, which is shown to improve classification performance in semi-supervised learning tasks on images. LMMVAE could be viewed as a method to add auxiliary LV, however its derivation is different and it is intended to improve unsupervised learning of large tabular datasets exhibiting complex correlation structures.

\subsection{Linear Mixed Models}
In developing LMMVAE we take inspiration from LMM, which we briefly describe and refer the reader to additional sources for a more thorough review. LMM make a distinction between the fixed and random parts of a model matrix, and their respective effects. In a typical linear regression setting, let $\mathbf{y}$ be a vector of $n$ observations we wish to relate to $\mathbf{X}$ as above, yet now $\mathbf{X}$ is treated as the fixed model matrix, holding $p$ ``global'' features which incur no dependence between observations. These are augmented by an additional $n \times q$ random model matrix $\mathbf{Z}$, holding $q$ features which are expected to induce correlation between observations, like time and location:
\begin{equation}\label{eq:classicLMM_lmmvae}
	\mathbf{y} = \mathbf{X}\bm{\beta} + \mathbf{Z}\mathbf{b} + \bm{\varepsilon}
\end{equation}
Here $\bm{\beta}$ is a $p$-dimensional vector of fixed model parameters or effects (FE), $\bm{\varepsilon}$ is normal i.i.d noise, and $\mathbf{b}$ is a $q$-dimensional vector of RE, which are random variables. Typically $\mathbf{b}$ is assumed to have a $\mathcal{N}(\mathbf{0}, \mathbf{D})$ distribution, where $\mathbf{D}$ is a $q \times q$ positive semi-definite matrix of appropriate structure, holding unknown {\em variance components} to be estimated. The structure of this covariance matrix remains unspecified to fit any kind of covariance scenario, but there are typically simplified structures used, for example a radial basis function (RBF) kernel for $q$ locations, a diagonal covariance with $\sigma^2_b$ on its diagonal for a single categorical feature with $q$ levels, and any combination of those. Accounting for this covariance structure is what gives LMM additional statistical power in comparison to ordinary linear regression, avoiding overfitted parameter estimates and ultimately improving prediction for unseen observations \cite{searle1992variance, McCulloch2008}.

A few papers have attempted to incorporate LMM into DNN, mostly for regression tasks on image and tabular datasets \cite{MeNets, DeepGLMM, LMMNN}. Ours is the first work that we are aware of, to take inspiration from LMM for dimensionality reduction with DNN.

\section{LMMVAE: Random Effects in VAE}
\label{sec:LMMVAE}
\subsection{Model and Architecture}
Looking back at \eqref{eq:PPCA}, suppose we are interested first in including a single categorical feature $v$ of cardinality $q$, in addition to the $p$ features of $\mathbf{x}_i$. One solution is one-hot encoding (OHE), where $q$ additional binary elements are added to $\mathbf{x}_i$, holding 1 for observation $i$'s level and 0 elsewhere. If $v$ is a high-cardinality feature like ``disease type'' for a group of patients, which could take thousands of levels -- this might lead to the overfitting of PPCA or even make it infeasible in extreme cases \cite{Hancock2020}. If it were a regression problem, we could take the LMM approach to include a high-cardinality categorical feature, for which we assume there is no need for a dedicated parameter for every level $j$. Rather, a RE vector $\mathbf{b}$ would be included in the model, coming from a $\mathcal{N}(\mathbf{0}, \sigma^2_b\mathbf{I}_q)$ distribution, enabling the inclusion of disease type in the model by a small $b_j$ addition for every disease $j$, with only a single variance parameter to estimate, $\sigma^2_b$.

Extending the linear relation between latent and output spaces to a general $f: \mathbb{R}^d \to \mathbb{R}^p$ as in \eqref{eq:VAE-as-PPCA}, the LMMVAE model for an additional high-cardinality categorical feature is therefore:
\begin{equation}
\label{eq:LMMVAE1}
    \mathbf{x}_{ij} = f(\mathbf{u}_{ij}) + \mathbf{b}_j + \bm{\varepsilon}_{ij}.
\end{equation}
$\mathbf{x}_{ij} \in \mathbb{R}^p$ is now the $i$-th observation in the $j$-th level of the categorical feature ($j = 1, \dots, q; i = 1, \dots, n_j$); $\mathbf{u}_{ij} \in \mathbb{R}^d$ is its LV with a $\mathcal{N}(\mathbf{0}, \mathbf{I}_d)$ prior as before; $\mathbf{b}_j \in \mathbb{R}^p$ is the $j$-th level RE addition to all $p$ ``fixed'' features; we omit the $\bm{\mu}$ term w.l.o.g as the features can be centered; and $\bm{\varepsilon}_{ij}$ Gaussian noise as before. $\mathbf{b}_j$ is sampled from a $\mathcal{N}(\mathbf{0}, \mathbf{D})$ distribution, where $\mathbf{D} = \text{diag}(\sigma^2_{b1}, \dots, \sigma^2_{bp})$ for a total of $p$ variance components, which adds flexibility to the model: a disease type could have a large effect on a feature like hospitalization time, but a small effect on a feature like age.

In order to be able to generalize the LMMVAE formulation to all covariance scenarios, we write \eqref{eq:LMMVAE1} in matrix form:
\begin{equation}
\label{eq:LMMVAE2}
    \mathbf{X} = f(\mathbf{U}) + \mathbf{Z}\mathbf{B} + \mathcal{E}.
\end{equation}
$\mathbf{X}$ is of order $n \times p$ as before; $\mathbf{U}$ of order $n \times d$ represents the LV stacked; $\mathbf{Z}$ is a $n \times q$ RE design matrix, which, in the case of a single categorical feature with cardinality $q$, is an indicator matrix holding 1 in its $ij$-th element if observation $i$ belongs to level $j$; $\mathbf{B}$ has the $\mathbf{b}_j$ RE vectors stacked into a $q \times p$ matrix; and $\mathcal{E}$ is the $n \times p$ Gaussian noise. $\mathbf{B}$ is in fact a sample from a matrix-normal distribution $\mathcal{MN}(\mathbf{0}, \mathbf{I}_q, \mathbf{D})$, where $\mathbf{D}$ is defined above. Some small details in this derivation may change for different covariance scenarios, but as we shall show the model in \eqref{eq:LMMVAE2} is general enough to include them all.

An important assumption LMMVAE makes, is that of independence between the ``fixed'' LV and RE given the data. In Bayesian terms we write $p_\theta(\mathbf{u}, \mathbf{b}|\mathbf{x}) = p_\theta(\mathbf{u}|\mathbf{x})p_\theta(\mathbf{b}|\mathbf{x})$. This assumption seems to work well with real datasets, simplifies LMMVAE's loss function and allows us to use two separate encoders, since if $q_\phi(\mathbf{u}, \mathbf{b}|\mathbf{x})$ is the surrogate model used to approximate $p_\theta(\mathbf{u}, \mathbf{b}|\mathbf{x})$, we assume it can also be factorized as:
\begin{equation}
\label{eq:independece}
    q_\phi(\mathbf{u}, \mathbf{b}|\mathbf{x}) = q_\phi(\mathbf{u}|\mathbf{x})q_\phi(\mathbf{b}|\mathbf{x})
\end{equation}
For both encoder models we use the same factorized Gaussian $\mathcal{N}(\bm{\mu}(x), \text{diag}(\bm{\sigma}^2(x)))$ as in VAE, one for the fixed LV $\mathbf{u}$ and one for the RE $\mathbf{b}$. We emphasize LMMVAE can be used with two separate encoders or a single encoder with double output -- FE output and RE output -- a perhaps simpler architecture which seems to be slightly inferior in terms of modeling quality (see Section~\ref{sec:experiments}).

\begin{figure}
    \centering
    \includegraphics[width=1.0\linewidth]{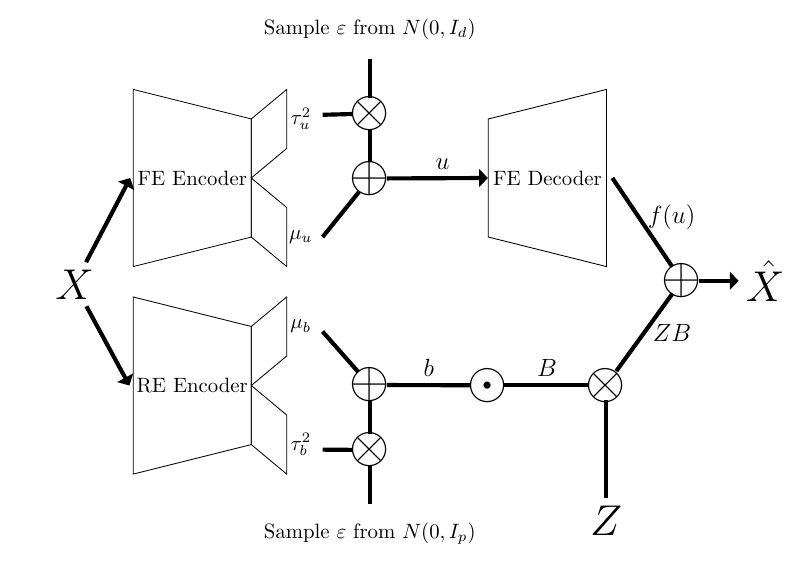}
    \caption{LMMVAE architecture: data $\mathbf{X}$ enters two separate FE and RE encoders (or a single encoder with double output), to produce the fixed LV $\mathbf{u}$ and RE $\mathbf{b}$ by the reparameterization trick. $\mathbf{u}$ goes through the FE decoder, its output $f(\mathbf{u})$ is added the RE term $\mathbf{Z}\mathbf{B}$ after $\mathbf{Z}$ enters the model and multiplies the RE matrix $\mathbf{B}$ after it had been properly formed from the $\mathbf{b}$ RE vectors, as depicted by the $\odot$ symbol (see the different covariance scenarios for more details). This produces the final reconstructions $\mathbf{\hat{X}}$.}
    \label{fig:lmmvae_scheme}
\end{figure}

A scheme for the LMMVAE architecture can be seen in Figure~\ref{fig:lmmvae_scheme}. For a given mini-batch, $\mathbf{X}$ enters into two separate encoders or a single encoder with double output. In this work we use MLP for tabular data and a convolutional neural network for image data: a FE encoder, producing the $d$ fixed means and variances $\bm{\mu}_u, \bm{\tau}^2_u$ with two final layers with $d$ neurons each, and a RE encoder, producing the $p$ random means and variances $\bm{\mu}_b, \bm{\tau}^2_b$ with two final layers of $p$ neurons each. In practice the actual variance parameter outputs are $\bm{\gamma}_u = \log \bm{\tau}^2_u$ and $\bm{\gamma}_b = \log \bm{\tau}^2_b$ to avoid putting positivity constraints, and these outputs are exponentiated when necessary. $\mathbf{u}$ and $\mathbf{b}$ are sampled with the reparameterization trick as in VAE, then $\mathbf{u}$ is passed through the FE decoder to reconstruct $\mathbf{X}$ in the output space, with a final layer of $p$ neurons. For each mini-batch of size $m$ we sample $m$ $p$-dimensional $\mathbf{b}$ vectors, but each of these samples belongs to a different level $j$ of the categorical feature (later in the longitudinal model this would be subject $j$ and in the spatial model location $j$). To get the $q \times p$ $\mathbf{B}$ matrix for a given mini-batch, we therefore average each level $j$'s $\mathbf{b}$ vectors, putting zeros for levels which are not present in the current mini-batch. This operation is depicted by the $\odot$ symbol in the scheme and is implemented in a vectorized fashion to speed up computations. $\mathbf{Z}$ is then input to the model, multiplied by $\mathbf{B}$, then the $\mathbf{Z}\mathbf{B}$ term is added to the FE encoder output $f(\mathbf{U})$ to produce the final output for the given mini-batch, $\mathbf{\hat{X}}$. Though not pursued here, we emphasize that the model in \eqref{eq:LMMVAE2} and the LMMVAE architecture can easily be modified to also include a decoder for the $\mathbf{Z}$ RE design matrix. Simply replace $\mathbf{Z}$ with $g_r(\mathbf{Z})$ and have $\mathbf{Z}$ go through yet another DNN to fit this relation.

Typically, one is interested in prediction for unseen data once the FE and RE encoders and the FE decoder have been learned. Let $(\mathbf{X}_{tr}, \mathbf{Z}_{tr})$ and $(\mathbf{X}_{te}, \mathbf{Z}_{te})$ be the training and testing datasets. For prediction of LV $\mathbf{U}_{te}$ on unseen data, one needs to run $\mathbf{X}_{te}$ through the learned FE encoder. For reconstruction of $\mathbf{\hat{X}}_{te}$, we first extract the full $q \times p$ estimate $\mathbf{\hat{B}}$ by running the learned RE encoder on the entire training data to get $n_{tr}$ vectors $\mathbf{b}$ and group them by the categorical feature's $q$ levels. The final reconstructed data is then $\mathbf{\hat{X}}_{te} = \hat{f}(\mathbf{U}_{te}) + \mathbf{Z}_{te}\mathbf{\hat{B}}$, where $\hat{f}$ is the learned FE decoder. Finally, exploring the $\mathbf{\hat{B}}$ matrix could be illuminating as to the categorical feature (and later temporal and spatial features) effects on the $p$ fixed features, as demonstrated in Section~\ref{sec:experiments}.

\subsection{Loss Function}
The assumption in \eqref{eq:independece} allows to decompose the KL divergence term in the ELBO in \eqref{eq:ELBO1} into fixed and random parts:
\begin{equation}
\label{eq:ELBO2}
\begin{aligned}
    \mathcal{L}_{\theta, \phi}(\mathbf{x}) =
    \mathbb{E}_{q_\phi(u, b|x)}[\log p_\theta(\mathbf{x}|\mathbf{u}, \mathbf{b})] - \beta \cdot D_{KL}[q_\phi(\mathbf{u}|\mathbf{x}) || p_\theta(\mathbf{u})] - \beta \cdot D_{KL}[q_\phi(\mathbf{b}|\mathbf{x}) || p_\theta(\mathbf{b})],
\end{aligned}
\end{equation}
where an extra hyperparameter $\beta$ multiplies the KL-divergence terms, as this $\beta$-VAE \cite{betaVAE} version proved to improve performance across most settings. See proof in Appendix~\ref{app:Dkl_decomposition}. The first and second terms are simplified as in ordinary VAE to be the squared reconstruction loss and the sum over the $d$ LV of a simple expression, respectively. All that is left is to simplify the third term, and this changes according to $\mathbf{B}$'s distribution. For the simple model described so far with a single high-cardinality categorical feature, where $\mathbf{B} \sim \mathcal{MN}(\mathbf{0}, \mathbf{I}_q, \mathbf{D})$ and $\mathbf{D} = \text{diag}(\sigma^2_{b1}, \dots, \sigma^2_{bp})$ one can show (Appendix~\ref{app:Loss}) that the final LMMVAE loss for a given observation $\mathbf{x}_{ij}$ is:
\begin{equation}
\label{eq:Loss0}
    \begin{aligned}
        \mathcal{L}_{\theta, \phi}(\mathbf{x}_{ij}) &= 
        (\mathbf{x}_{ij}-\mathbf{\hat{x}}_{ij})^2
        -\frac{\beta}{2}\sum_{l=1}^d\left[1 + \gamma_{ul} - e^{\gamma_{ul}} - \mu^2_{ul}\right] \\
        &-\frac{\beta}{2}\sum_{k=1}^{p}\left[1 + \gamma_{bk} -\delta_{bk} - e^{\gamma_{bk} - \delta_{bk}} - \mu^2_{bk}e^{-\delta_{bk}}\right]
    \end{aligned}
\end{equation}
Here $\bm{\delta}_b = \log \bm{\sigma}^2_b$, the log of the prior RE variance terms. One could either treat them as known priors and specify them, tune them via grid search or fit them via SGD. We take the prior approach, assume during training that $\mathbf{D} = \sigma^2_b\mathbf{I}_p$ and specify $\sigma^2_b = 1$ (or $\delta_b = 0$) always, and demonstrate the robustness of LMMVAE when this assumption is false.

\subsection{Common Covariance Scenarios}
\label{sec:LMMVAE:scenarios}
In this Section we derive solutions to common correlation scenarios such as high-cardinality categorical data, longitudinal and spatial data.

\subsubsection{High-Cardinality Categorical Data}
\label{sec:LMMVAE:categorical}
Suppose there are $K$ additional high-cardinality categorical features, each having $q_k$ cardinality, and its own $\mathbf{B}_k$ matrix of order $q_k \times p$, distributed $\mathcal{MN}(\mathbf{0}, \mathbf{I}_{q_k}, \mathbf{D}_k)$, and $\mathbf{D}_k$ is diagonal with $p$ variance components as before, hence we assume these RE features are uncorrelated. Model \eqref{eq:LMMVAE2} requires minor changes: $\mathbf{Z}$ is now of order $n \times Q$, where $Q = \sum_k q_k$, it can be viewed as horizontal stacking of $K$ indicator matrices: $\mathbf{Z} = [\mathbf{Z}_1 \vdots \dots \vdots \mathbf{Z}_K]$. The $\mathbf{B}_k$ RE matrices are vertically stacked to form $\mathbf{B}$ of order $Q \times p$.

The architecture in Figure~\ref{fig:lmmvae_scheme} changes so that the RE encoder outputs $K$ pairs of means and (log) variances, one for each categorical feature. This means we fit a different encoder $q_\phi(\mathbf{b}_k|\mathbf{x})$ for each of the $K$ features, with some parameters shared. From these outputs the $\mathbf{B}_k$s are sampled as before and multiplied by the $\mathbf{Z}_k$s as before, to form the $\mathbf{Z}_k\mathbf{B}_k$ terms which are added to the $f(\mathbf{u})$ FE decoder output to form the final reconstruction. Finally, the third term of the loss in \eqref{eq:Loss0} is replaced with a sum over $K$ KL-divergence terms: $-\frac{\beta}{2}\sum_{k=1}^{K}D_{KL}[q_\phi(\mathbf{b}_k|\mathbf{x}) || p_\theta(\mathbf{b}_k)]$.

\subsubsection{Longitudinal Data}
\label{sec:LMMVAE:longitudinal}
Typically in a longitudinal dataset $q$ unique units of measurements (e.g. human subjects) are repeatedly measured through time $t$, inducing temporal correlation between observations. Let $\mathbf{x}_{ij} \in \mathbb{R}^p$ be the measurement of subject $j$ at time $t_{ij}$, where $i = 1, \dots, n_j$, i.e. the number of measurements and times are allowed to vary between subjects. The LMMVAE longitudinal model assumes a random addition for the $p$ fixed features for each subject (``random intercept'') $\mathbf{b}_{0j} \in \mathbb{R}^p$, a random addition for each subject which is linear with time $t$ (``random slope'') $t_{ij} \cdot \mathbf{b}_{1j} \in \mathbb{R}^p$, and so on until polynomial order of $K - 1$. For $K = 3$ terms for instance, the LMMVAE model in vector form is:
\begin{equation}
\label{eq:LMMVAElongitudinal}
    \mathbf{x}_{ij} = f(\mathbf{u}_{ij}) + \mathbf{b}_{0j} + t_{ij} \cdot \mathbf{b}_{1j} + t^2_{ij} \cdot \mathbf{b}_{2j} + \bm{\varepsilon}_{ij}.
\end{equation}
Notice each of the $\mathbf{b}_{kj}$ ($k = 0, \dots, K - 1$) is a vector while $t_{ij}$ is a scalar. To keep the model in matrix form in \eqref{eq:LMMVAE2} intact, the $\mathbf{b}_{kj}$ are vertically stacked to form the $\mathbf{B}$ matrix of order $Kq \times p$ which is distributed $\mathcal{MN}(\mathbf{0}, \bm{\Phi} \otimes \mathbf{I}_{q}, \mathbf{D})$, where $\bm{\Phi}$ of order $K \times K$ holds the covariance structure of the $K$ polynomial terms. To form the general $\mathbf{Z}$ matrix: assume $\mathbf{t}$ is the full $n$-length vector of measurement times, and note $\mathbf{t}$ can have identical entries if different subjects were measured at the same time. Let $\mathbf{Z}_0$ be the $n \times q$ binary matrix where the $\left[l, j\right]$-th entry equals 1 if row $l$ is a measurement of subject $j$ at the time indicated by $t_l$. The full $\mathbf{Z}$ would be of order $n \times Kq$ for $K$ polynomial terms and $q$ subjects. $\mathbf{Z}$ would be the horizontal stacking of $K$ matrices: $[\mathbf{Z}_0 \vdots \dots \vdots \mathbf{Z}_{K - 1}]$ where each $\mathbf{Z}_k = diag(\mathbf{t}^{k}) \cdot \mathbf{Z}_0$ for $k = 0, \dots, K - 1$.

In terms of architecture and loss this is very similar to the case of multiple high-cardinality categorical features discussed above. The RE encoder outputs $K$ pairs of means and (log) variances, essentially making $K$ encoders $q_\phi(\mathbf{b}_k|\mathbf{x})$ for each of the $K$ polynomial terms, from which the $\mathbf{B}_k$'s are sampled, stacked to form $\mathbf{B}$ and multiplied by the $\mathbf{Z}$ input. The change to the loss term is also similar, the only difference is the sum over KL-divergence terms running from $k = 0$ to $K - 1$, to keep with the longitudinal notation.

\subsubsection{Spatial Data}
\label{sec:LMMVAE:spatial}
Many datasets contain a set of longitude/latitude features, pointing to measurements taken at $q$ unique locations, which means observations might be spatially correlated. We expect observations to be similar to each other as distance between them gets smaller, and one approach to capture such correlations is through the Gaussian process framework or kriging \cite{Rasmussen, Cressie1993}. The LMMVAE formulation for the $i$-th observation in the $j$-th location remains the same as in ~\eqref{eq:LMMVAE1} and \eqref{eq:LMMVAE2} for the single categorical feature: we expect a $\mathbf{b}_j \in \mathbb{R}^p$ addition to the $p$ fixed features from location $j$, and stack the vectors $\mathbf{b}_1, \dots, \mathbf{b}_q$ to form the $q \times p$ RE matrix $\mathbf{B}$. However, now $\mathbf{B}$ has distribution $\mathcal{MN}(\mathbf{0}, \mathbf{K}, \mathbf{D})$, where $\mathbf{K}$ is the $q \times q$ kernel matrix, reflecting the correlations between $q$ locations (rows), where in this work we use the RBF kernel always. If $\mathbf{s}_j, \mathbf{s}_{j'}$ are two locations, the $\left[j, j'\right]$-th entry of $\mathbf{K}$ is $k(\mathbf{s}_j, \mathbf{s}_{j'}) = \exp \frac{-\|\mathbf{s}_j - \mathbf{s}_{j'}\|^2}{2l^2}$, where $l^2$ can be seen as another variance parameter. Furthermore, $\mathbf{D} = \text{diag}(\sigma^2_{b1}, \dots, \sigma^2_{bp})$ as before, and $\mathbf{Z}$ remains the $n \times q$ indicator matrix pointing to the $j$-th location of every row.

The change of architecture and loss for the spatial model is less trivial, since this model assumes a prior between-rows correlation of the RE $\mathbf{B}$ matrix: each of $\mathbf{B}$'s \emph{columns} $\mathbf{b}_{k} \in \mathbb{R}^q$ is distributed $\mathcal{N}(\mathbf{0}, \mathbf{K})$. A well known result from LMM is that the posterior distribution of $\mathbf{b}$ given data $\mathbf{y}$ structured in \eqref{eq:classicLMM_lmmvae} with $\mathbf{b} \sim \mathcal{N}(\mathbf{0}, \mathbf{K})$ is $\mathcal{N}(\bm{\mu}^*, \bm{\Psi}^*)$, where $\bm{\mu}^* = \mathbf{K}\mathbf{Z}^T\mathbf{V}^{-1}(\mathbf{y} - \mathbf{X}\bm{\beta})$ is the best linear unbiased predictor (BLUP) of $\mathbf{b}$, $\mathbf{V} = \mathbf{Z}\mathbf{K}\mathbf{Z}^T + \sigma^2_e\mathbf{I}$ is the marginal covariance of $\mathbf{y}$ and $\bm{\Psi}^* = \mathbf{K} - \mathbf{K}\mathbf{Z}^T\mathbf{V}^{-1}\mathbf{Z}\mathbf{K}$ \cite{McCulloch2008}. In the spatial model we found it useful to assume $\mathbf{K}$ \emph{known} with a constant $l^2 = 1$ parameter, then compute $\bm{\Psi}^*$ once on the training set and its Cholesky decomposition $\bm{\Psi}^* = \bm{\Psi}^{*\frac{1}{2}}\bm{\Psi}^{T*\frac{1}{2}}$. With the current LMMVAE architecture one could think of $\mathbf{B}$'s posterior for a single categorical feature as $\mathcal{MN}(\mathbf{0}, \mathbf{I}_q, \mathbf{D})$, and so by multiplying it on the left by $\bm{\Psi}^{*\frac{1}{2}}$ for each mini-batch, we enforce the desired between-rows correlation on its structure \cite{gupta2018matrix}. The loss function, therefore, remains the same as in \eqref{eq:Loss0}. We note that in order to compute $\bm{\Psi}^*$ there is a need for inverting and storing $\mathbf{V}$ which is of order $n \times n$. In practice we found that uniformly sampling $n_{samp} = 10000$ of $\mathbf{Z}$'s rows works well, and one could use more sophisticated approaches such as inducing points \cite{JMLR:v6:quinonero-candela05a}.

\section{Experiments}
\label{sec:experiments}
In this Section we report our experiments with LMMVAE on simulated and real data, focusing on large tabular datasets, and on reconstruction loss of the $p$ fixed features on {\em unseen} data as well as NLL, as the main metrics to minimize. We summarize key results and refer to Appendices ~\ref{app:additional} and ~\ref{app:additional:real} for additional results and plots. All experiments were implemented in Python using Keras \cite{chollet2015keras} and Tensorflow \cite{tensorflow2015-whitepaper}, run on Google Colab with NVIDIA Tesla T4 GPU machines, the code is available on Github at \url{https://github.com/gsimchoni/lmmvae}.

\subsection{Simulated Data}
\label{sec:experiments:simulated}
We simulate separately for the high-cardinality, longitudinal 
 and spatial scenarios. In all simulations, we sample LV $\mathbf{U}$ of order $n \times d$ from a $\mathcal{N}(0, 1)$ distribution, where $n = 100000$ always, and $d$ is varied in $\{1, 2\}$, as is commonly required for tabular datasets (see Section~\ref{sec:results:real} for higher $d$s). To map the LV to output space of $p = 100$ fixed features, a helper matrix $\mathbf{W}$ of order $p \times d$ is sampled from a $\mathcal{N}(0, 1)$ as well. The non-linear $f: \mathbb{R}^{d} \to \mathbb{R}^{p}$ function used is $(\mathbf{u}_l \cdot \mathbf{W}^T) * \cos(\mathbf{u}_l \cdot \mathbf{W}^T)$, where $\mathbf{u}_l$ is the $l$-th row of $\mathbf{U}$, $*$ is elementwise multiplication and $f$ is applied rowwise. The final matrix $\mathbf{X}$ of order $n \times p$ following model \eqref{eq:LMMVAE2} is:
\begin{equation}
    \label{eq:f_lmmvae}
    \mathbf{X} = f(\mathbf{U}) + \mathcal{M} + \mathbf{Z}\mathbf{B} + \mathcal{E},
\end{equation}
where $\mathcal{M}$ of order $n \times p$ is a vector of feature means $\bm{\mu} \in \mathbb{R}^p$ replicated $n$ times for all observations and sampled from $\mathcal{U}(-10, 10)$, and $\mathcal{E}$ of order $n \times p$ is $\mathcal{N}(0, 1)$ noise. As described in Section~\ref{sec:LMMVAE:scenarios}, the only element which changes from one covariance scenario to another is $\mathbf{Z}\mathbf{B}$, see next. Common to all scenarios however is the need to sample $q$ groups with different $n_j$s summing to $n$ (levels of a categorical feature, subjects in a longitudinal study, locations). The $q$ levels $n_j$s are generated using multinomial distribution sampling. For each combination of $d$ and the various variance parameters we perform 5 iterations in which we sample the data, randomly split it into training (80\%) $(\mathbf{X}_{tr}, \mathbf{Z}_{tr})$ and testing (20\%) $(\mathbf{X}_{te}, \mathbf{Z}_{te})$, train our models on $(\mathbf{X}_{tr}, \mathbf{Z}_{tr})$ to predict $\mathbf{\hat{X}}_{te}$ and compare the bottom-line test squared reconstruction error and NLL loss. We use the same DNN architecture for all encoders, that is an MLP with 2 hidden layers with $\{1000, 500\}$ neurons, a ReLU activation and a final output layer with $d$ neurons, hence $d$ is assumed known. The decoders architecture is similar, in reverse. A batch size of 1000 is used, with 200 epochs. Finally a prior of 1 is used for all $\sigma^2$ variance parameters and $\beta = 0.01$. We did not systematically examine the effect of these parameters on the results, so we believe LMMVAE's results could even be improved with a proper search on these hyperparameters.

\subsubsection{High-Cardinality Categorical Data}
\label{sec:experiments:simulated:categorical}
\begin{figure}
    \centering
    \includegraphics[width=1.0\linewidth]{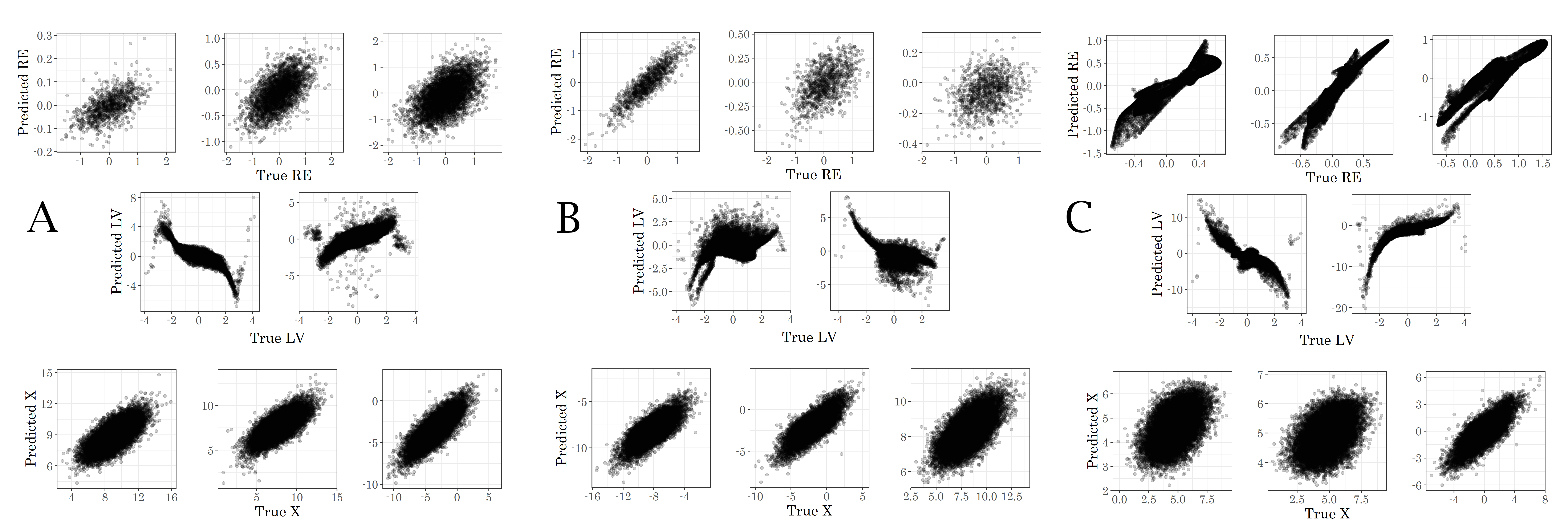}
    \caption{Predicted vs. true scatter plots for simulated datasets with $n = 100000$ observations. First row: first column of $\mathbf{B}_1, B_2, B_3$. Second row: LV $U$ (here $d = 2$). Third row: first 3 columns of $\mathbf{X}_{te}$. A: Three high-cardinality categorical features, with $q_1 = 1000, q_2 = 3000, q_3 = 5000$. B: Longitudinal data with $q = 1000$ subjects, and $K = 3$ polynomial terms on $t$, random mode. C: spatial data with $q = 10000$ locations, and an RBF kernel.}
    \label{fig:simulated}
\end{figure}

$K = 3$ categorical features are used. Their cardinality is $q_1 = 1000, q_2 = 3000, q_3 = 5000$, so $\mathbf{Z}$ is of order $100000 \times 9000$ and $\mathbf{B}$ of order $9000 \times 100$. The $p$ variance terms on $\mathbf{D}_k$ for each categorical feature $k$ are generated using $[\text{Poisson}(\sigma^2_{bk}) + 1]\cdot c$ sampling with $c = \min(\sigma^2_{bk}, 1)$, and those $[\sigma^2_{b1}, \sigma^2_{b2}, \sigma^2_{b3}]$ are varied in $\{0.3, 3.0\}$. Here we compare LMMVAE to PCA with OHE on the categorical features and PCA ignoring them altogether, and to VAE with OHE, VAE with entity embeddings on the categorical features and VAE ignoring them. We also compare LMMVAE to ``LMMVAE-I'' which uses a single encoder with double output. For brevity we show in Table~\ref{tab:simulated:all} results for when $\sigma^2_{bk} = 0.3$ for all $k$ and refer the interested reader to Appendix~\ref{app:additional} where the rest of the results can be found, as well as mean runtimes and NLL loss. Finally, we show in Figure~\ref{fig:simulated} predicted-versus-true scatter plots for the first columns of $\mathbf{B}$, $\mathbf{U}_{te}$ and $\mathbf{X}_{te}$. LMMVAE is able to not only reconstruct $\mathbf{X}_{te}$ well, but also its different components, the LV and RE.

As can be seen, LMMVAE's reconstruction and NLL losses are superior to all other methods, with a reasonable cost in runtime. It is further interesting to see how VAE performs uniformly better than PCA, and how using OHE with cardinality as high as in this experiment harms reconstruction loss and NLL in comparison to ignoring the categorical features.

\begin{table}
  \caption{Mean test reconstruction errors and NLL loss for some VAE methods for simulated models. Left: 3 categorical features, with $q_1 = 1000, q_2 = 3000, q_3 = 5000$. Middle: longitudinal data with $q = 1000$ subjects, and $K = 3$ polynomial terms on $t$. Right: spatial data with $q = 10000$ locations, and an RBF kernel. Standard errors in parentheses, bold results are non-inferior to the best result in a paired t-test. Included here only datasets for which all variance components are 0.3, for additional datasets and mean running times see Appendix.}
  \label{tab:simulated:all}
  \begin{adjustbox}{width=\columnwidth,center}
  \centering
  \begin{tabular}{l|ll|ll|ll|ll|ll|}
\cline{2-11}
& \multicolumn{2}{c|}{\textbf{Categorical Data}} & \multicolumn{4}{c|}{\textbf{Longitudinal Data}} & \multicolumn{4}{c|}{\textbf{Spatial Data}}\\
\cline{2-11}
 & & & \multicolumn{2}{c|}{\textbf{Random mode}} & \multicolumn{2}{c|}{\textbf{Future mode}} & \multicolumn{2}{c|}{\textbf{Random mode}} & \multicolumn{2}{c|}{\textbf{Unknown mode}}\\
\cline{2-11}
 & $d = 1$ & $d = 2$ & $d = 1$ & $d = 2$ & $d = 1$ & $d = 2$ & $d = 1$ & $d = 2$ & $d = 1$ & $d = 2$ \\
    \midrule
    & \multicolumn{10}{c|}{\textbf{Mean test reconstruction err.}} \\
    \midrule
    PCA-Ignore & 2.31 (.02) & 2.56 (.02) & 1.78 (.02) & 1.96 (.02) & 2.26 (.03) & 2.39 (.03) & 1.49 (.02) & 1.71 (.05) & 1.46 (.02)  & 1.68 (.03) \\
    PCA-OHE & 2.31 (.02) & 2.58 (.02) & 1.82 (.02) & 1.98 (.02) & 2.22 (.01) & 2.41 (.04) & 1.47 (.02) & 1.66 (.03) & 1.48 (.03)  & 1.67 (.04) \\
    VAE-Ignore & 2.15 (.01) & 2.21 (.03) & 1.60 (.00) & 1.63 (.00) & 2.09 (.03) & 2.06 (.01) & 1.31 (.05) & 1.28 (.04) & 1.23 (.01)  & 1.33 (.04) \\
    VAE-OHE & 2.24 (.01) & 2.32 (.02) & -- & -- & -- & -- & -- & --& -- & -- \\
    VAE-Embed. & 2.16 (.01) & 2.26 (.03) & 1.61 (.00) & 1.65 (.02) & 2.07 (.01) & 2.13 (.02) & 1.31 (.01) & 1.41 (.04) & 1.31 (.01)  & 1.35 (.01) \\
    SVGPVAE & -- & -- & 2.10 (.0) & 2.46 (.0) & 2.49 (.03) & 2.90 (.03) & 1.58 (.05) & 1.92 (.02) & 1.85 (.03)  & 2.36 (.04) \\
    VRAE & -- & -- & 2.06 (.0) & 2.54 (.0) & 2.48 (.01) & 2.94 (.06) & -- & -- & -- & -- \\
    LMMVAE-I & \textbf{1.34 (.02)} & \textbf{1.47 (.03)} & \textbf{1.11 (.01)} & 1.18 (.02) & \textbf{1.24 (.01)}  & 1.30 (.02) & -- & -- & -- & -- \\
    LMMVAE & \textbf{1.28 (.02)} & \textbf{1.52 (.04)} & \textbf{1.11 (.00)} & \textbf{1.11 (.02)} & \textbf{1.22 (.02)}  & \textbf{1.22 (.01)} & \textbf{1.02 (.00)} & \textbf{1.10 (.03)} & \textbf{1.04 (.01)}  & \textbf{1.10 (.03)} \\
    \midrule
    & \multicolumn{10}{c|}{\textbf{Mean test NLL loss}} \\
    \midrule
    VAE-Ignore & 214.6 & 221.5 & 160.0 & 162.9 & 209.3 & 206.3 & \textbf{131.6} & \textbf{133.9} & \textbf{128.8} & 135.8 \\
    VAE-Embed. & 216.5 & 225.8 & 160.6 & 165.5 & 207.2 & 213.3 & \textbf{131.6} & \textbf{131.8} & 141.8 & 143.7 \\
    LMMVAE & \textbf{96.7} & \textbf{113.3} & \textbf{106.1} & \textbf{106.8} & \textbf{99.3} & \textbf{99.4} & \textbf{122.0} & \textbf{128.2} & \textbf{131.8} & \textbf{133.3} \\
\bottomrule
\end{tabular}

  \end{adjustbox}
\end{table}

\subsubsection{Longitudinal Data}
\label{sec:experiments:simulated:longitudinal}
A model identical to \eqref{eq:LMMVAElongitudinal} is used with $K = 3$ polynomial terms, where $f(\mathbf{u}_{ij})$ is the same as in the categorical data experiment. As before, a variable number of $n_j$ measurements is sampled for each of $q = 1000$ subjects, the total number of observations being $n = 100000$. For each subject $t$ is sampled from a $\mathcal{U}(0, 1)$ distribution. $\mathbf{B}$ is sampled from $\mathcal{MN}(\mathbf{0}, \bm{\Phi} \otimes \mathbf{I}_{q}, \mathbf{D})$ as described, where $\mathbf{D}$ is $\mathbf{I}_p$ always, and $\bm{\Phi}$ is varied as follows. We vary $[\sigma^2_{b_0}, \sigma^2_{b_1}, \sigma^2_{b_2}]$ on $\bm{\Phi}$'s diagonal in $\{0.3, 3.0\}$, and since the current LMMVAE architecture does not explicitly take into account possible correlations between polynomial terms, we challenge it by adding a covariance of 0.3 between two of the possible three off-diagonal relations in $\bm{\Phi}$: between the intercept and slope terms and between the intercept and quadratic terms. Furthermore, to make the simulation more realistic we not only include a ``Random'' mode where the data is split randomly to 80\% training and 20\% testing sets, but also a ``Future'' mode where the  testing set are those 20\% observations which occur latest in time $t$ across all $n$ observations, meaning that the model is only trained on past observations. We compare LMMVAE's results to PCA and VAE as before but also to SVGPVAE and VRAE with a LSTM layer with 100 neurons, chosen after tuning.

Table~\ref{tab:simulated:all} shows results for $\sigma^2_k = 0.3$ for all $K$ polynomial terms, where the rest of the results can be found in Appendix~\ref{app:additional}. Clearly, LMMVAE's performance is superior to all methods in both modes. See also Figure~\ref{fig:simulated} for predicted-versus-true scatter plots for $\mathbf{B}$, $\mathbf{U}_{te}$ and $\mathbf{X}_{te}$.

\subsubsection{Spatial Data}
\label{sec:experiments:simulated:spatial}
$q = 10000$ 2-D locations are sampled on the $\mathcal{U}[-1, 1] \times \mathcal{U}[-1, 1]$ grid, the total number of observations being $n = 100000$ as before. $\mathbf{B}$ is sampled from $\mathcal{MN}(\mathbf{0}, \mathbf{K}, \mathbf{D})$ as described, where $\mathbf{K}$ is a standard RBF kernel with parameter $l^2$ varied in $\{0.3, 3.0\}$, and the $p$ variance terms on $\mathbf{D}$'s diagonal are generated using Poisson sampling, so that they are not identical, but concentrated around a mean $\sigma^2_{b}$ which is varied in $\{0.3, 3.0\}$ as well. In addition to a ``Random'' mode where the data is split randomly to 80\% training and 20\% testing sets, we also include an ``Unknown'' mode in which the testing set is comprised entirely from unseen locations. Table~\ref{tab:simulated:all} shows results for  $\sigma^2_b = l^2 = 0.3$, the rest of the results can be seen in Appendix~\ref{app:additional}, where LMMVAE consistently outperforms other approaches. We also show in Figure~\ref{fig:simulated} predicted versus true scatter plots of the RE features, LV and columns of $\mathbf{X}_{te}$.

\subsection{Real Data}
\label{sec:results:real}
\begin{figure}
    \centering
    \includegraphics[width=0.85\linewidth]{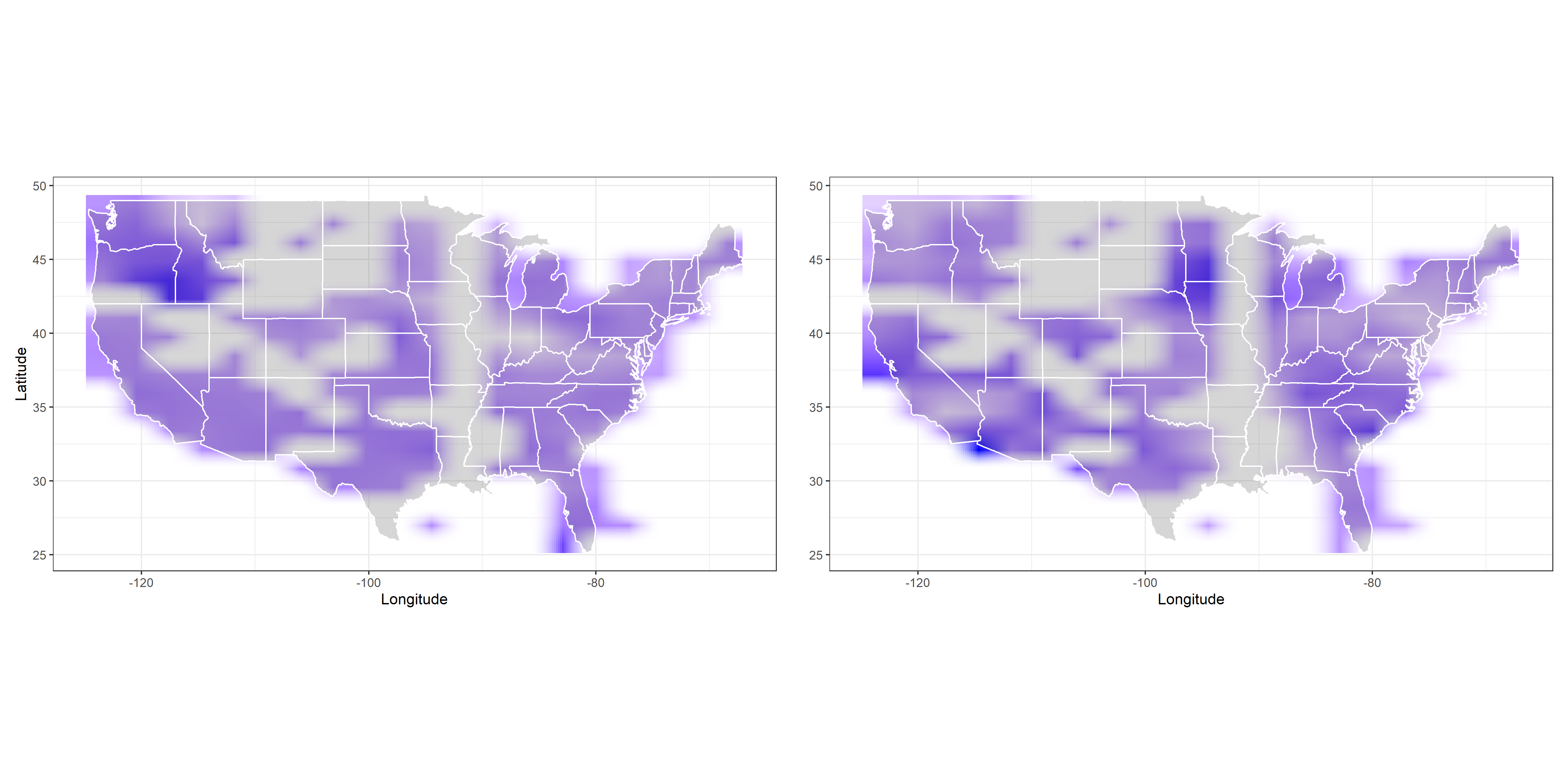}
    \caption{Exploring the $\mathbf{\hat{B}}$ RE matrix from the Cars dataset containing spatial features, with a raster plot. Left: distribution across the US of the $\mathbf{\hat{B}}$ column corresponding to the price feature; Right: distribution across the US of the $\mathbf{\hat{B}}$ column corresponding to the odometer feature.}
    \label{fig:real:B_hat}
\end{figure}

\begin{figure}
    \centering
    \includegraphics[width=0.85\linewidth]{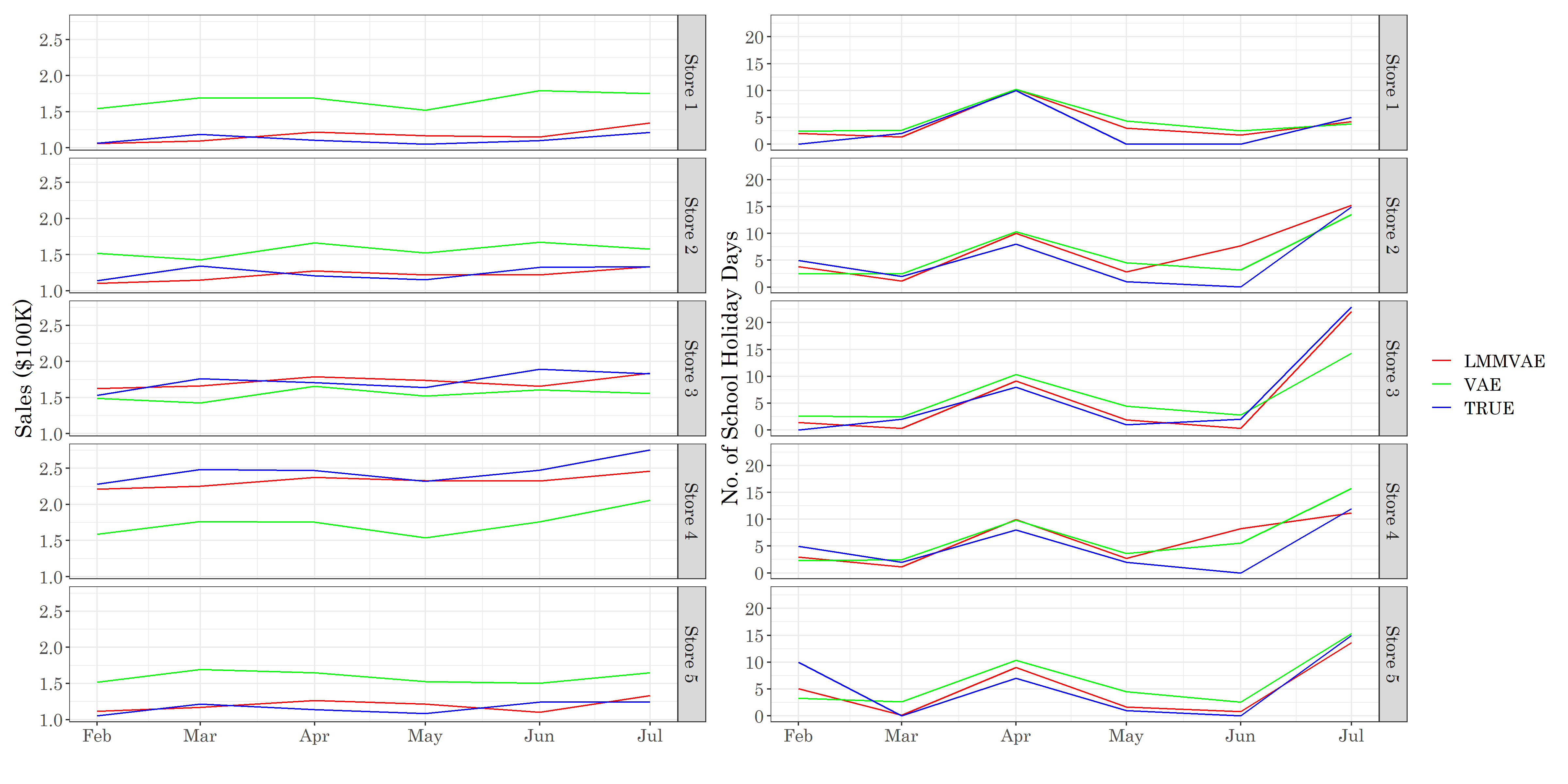}
    \caption{Comparing true vs. reconstructed $\mathbf{X}_{te}$ for the Rossmann stores longitudinal dataset, Future mode with $d = 1$. In this mode the model is trained on the first 25 months of dataset to reconstruct the last 6 months of the dataset. Left: comparing LMMVAE and VAE with entity embeddings on the sales feature; Right: comparing LMMVAE and VAE with entity embeddings on the school days feature.}
    \label{fig:real:X_te_recon}
\end{figure}

We used two real datasets for each covariance scenario, their full details can be found in Appendix~\ref{app:realdetails}, and some technical details can be found in Table~\ref{app:tab:real-data-details2}. Additionally we used two real datasets demonstrating a combination of the categorical and spatial covariance scenarios, for example the Cars dataset which contains details regarding 97K cars and has spatial features (12K longitude/latitude locations across the US) as well as a high-cardinality categorical feature (the car's model, with 15K models). A 5-CV procedure was carried out with DNN architecture, number of epochs and batch size identical to simulations, and $d$ was varied in $\{1, 2, 5\}$. The $\beta$ and $\sigma^2_b$ prior were chosen through tuning, and for the longitudinal datasets we used $K = 2$ polynomial terms. The mean test reconstruction error for $d = 2$ and Random mode can be found in Table~\ref{tab:real_mse_short}. In Appendix~\ref{app:additional:real} we provide results for additional $d$s, Future/Unknown modes, and comparisons to VRAE and SVGPVAE where relevant. LMMVAE's performance on real datasets is clearly the best with the mean reconstruction error and NLL losses gaps being considerably high for the longitudinal and spatial scenarios. We supply two visualizations to demonstrate LMMVAE's strengths: in Figure~\ref{fig:real:B_hat} we show the $\mathbf{\hat{B}}$ RE matrix for two of the features of the Cars dataset across the US map. This kind of plots can be used by practitioners to further explore interesting patterns in the data, in similar ways to how predicted RE have been used in data analysis in LMM \cite{McCulloch2008}. In Figure~\ref{fig:real:X_te_recon} we further show the reconstructed $\mathbf{X}_{te}$ matrix (two features) for the first five stores from the Rossmann longitudinal dataset, in Future mode with $d = 1$, comparing LMMVAE to VAE with entity embeddings. As can be seen LMMVAE's forecasting of the last unseen six months data is equivalent or considerably better than VAE, for these two features.

\begin{table}
  \caption{Real datasets specifics and training details.}
  \label{app:tab:real-data-details2}
  \centering
  \begin{tabular}{l|lllllllll}
    \toprule
    Dataset & $n$ & $n_j$ & $p$ & $q$ & neurons & batch & epochs & $\beta$ & $\sigma^2_b$ prior \\
    \midrule
    News & 81K & -- & 176 & {\small 5K, 72K} & [1000, 500] & 1000 & 200 & 0.01 & 0.01 \\
    Spotify & 28K & -- & 14 & \makecell{\small 10K, 22K \\ \small 2K, 0.5K} & [1000, 500] & 1000 & 200 & 0.01 & 0.01 \\
    Rossmann & 33K & 25-31 & 23 & 1K & [1000, 500] & 1000 & 200 & 0.01 & 1.0 \\
    UKB & 528K & 1-4 & 50 & 469K & [1000, 500] & 1000 & 200 & 0.01 & 0.001 \\
    Income & 71K & 1-2K & 30 & 3K & [1000, 500] & 1000 & 200 & 0.01 & 0.01 \\
    Asthma & 69K & 1-2K & 31 & 3K & [1000, 500] & 1000 & 200 & 0.01 & 0.1 \\
    Cars & 97K & 1-632 & 73 & {\small 12K, 15K} & [1000, 500] & 1000 & 200 & 0.01 & 0.01 \\
    Airbnb & 50K & 1-404 & 196 & {\small 3K, 40K} & [1000, 500] & 1000 & 200 & 0.01 & 0.01 \\
    CelebA & 202K & 1-35 & \small{72x60x3} & 10K & [32, 16]\small{(Conv)} & 100 & 50 & 0.01 & 0.001 \\
\end{tabular}
\end{table}

\begin{table}
  \caption{Mean test reconstruction errors for real datasets with latent dimension $d = 2$, over 5-CV runs, random mode. For the CelebA images dataset $d = 100$ is used.}
  \label{tab:real_mse_short}
  \centering
  \begin{tabular}{l|llll|llll}
\toprule
Dataset & $n$ & $n_j$ & $p$ & $q$ & PCA-Ig. & VAE-Ig. & VAE-Em. & LMMVAE \\
    \midrule
    \multicolumn{9}{c}{\textbf{Categorical}} \\
	\midrule
    News & 81K & -- & 176 & {\small 5K, 72K} &  .95 (.00) & .68 (.00) & .75 (.00) & \textbf{.64 (.00)} \\
    Spotify & 28K & -- & 14 & \makecell{\small 10K, 22K \\ \small 2K, 0.5K} & .74 (.00) & \textbf{.49 (.01)} & .69 (.00) & \textbf{.48 (.01)} \\
	\midrule
	\multicolumn{9}{c}{\textbf{Longitudinal}} \\
	\midrule
    Rossm. & 33K & 25-31 & 23 & 1K &  .78 (.01) & .20 (.01) & .21 (.01) & \textbf{.01 (.00)} \\
    UKB & 528K & 1-4 & 50 & 469K & .87 (.00) & .64 (.00) & .75 (.00) & \textbf{.63 (.00)}  \\
    \midrule
    \multicolumn{9}{c}{\textbf{Spatial}} \\
	\midrule
    Income & 71K & 1-2K & 30 & 3K & .014 (.00) & .012 (.00) & .011 (.00) & \textbf{.004 (.00)} \\
    Asthma & 69K & 1-2K & 31 & 3K & .015 (.00) & .016 (.00) & .016 (.00) & \textbf{.006 (.00)} \\
    \midrule
    \multicolumn{9}{c}{\textbf{Spatial-Categorical combination}} \\
	\midrule
    Cars & 97K & 1-632 & 73 & {\small 12K, 15K} & .081 (.00) & .065 (.00) & .068 (.00) & \textbf{.028 (.00)}  \\
    Airbnb & 50K & 1-404 & 196 & {\small 3K, 40K} & .061 (.00) & .053 (.00) & .052 (.00) & \textbf{.050 (.00)}  \\
    \midrule
    \multicolumn{9}{c}{\textbf{Image Data}} \\
	\midrule
    CelebA & 202K & 1-35 & 13K & 10K & -- & \textbf{.009 (.00)} & .051 (.00) & \textbf{.009 (.00)}  \\
\bottomrule
\end{tabular}
\end{table}

In addition to tabular datasets we tested our approach on a large image dataset -- the CelebA dataset \cite{liu2015faceattributes}, comprising of 202K facial images belonging to 10,177 individuals, which can be thought of as a single high-cardinality categorical feature \cite{lmmnn_neurips}. Table~\ref{tab:real_mse_short} shows mean test reconstruction error for $d = 100$, where convolutional encoders/decoders are used for all methods. While LMMVAE did not improve over VAE ignoring the face identity in this case,  it demonstrates LMMVAE's scalability to higher $d$s which are typically required for image data. Given success on tabular data, we expect that LMMVAE will be more impactful for image data with additional external features, but more research is needed. Additional $d$s can be found in Appendix~\ref{app:additional:real}, as well as test faces reconstructed with LMMVAE.

\subsection{Downstream Tasks on Real Data Latent Representations}
\label{sec:results:downstream}
Along with examining the NLL loss and reconstruction error for unseen datasets we also conducted a set of experiments in which the latent representations generated by LMMVAE were tested in a downstream supervised classification task. Performing dimensionality reduction brings us to the $\mathbf{\hat{U}}_{tr}, \mathbf{\hat{U}}_{te}$ LV with dimensions $n_{tr} \times d$ and $n_{te} \times d$ respectively. We then take some low-dimensional categorical feature $\mathbf{y}_{tr}, \mathbf{y}_{te}$ (not used in dimensionality reduction), train a multi-class classifier on $(\mathbf{\hat{U}}_{tr}, \mathbf{y}_{tr})$ and predict on $(\mathbf{\hat{U}}_{te}, \mathbf{y}_{te})$. Table~\ref{tab:downstream_cars} summarizes the 5-CV mean test micro-average AUC for a standard $k$-NN classifier with $k = 100$ neighbors, where for two out of three datasets dimensionality reduction has clearly improved classification performance, and for small $d$ LMMVAE outperforms the other methods.

\begin{table}
    \caption{Mean 5-CV test micro-average AUC (higher is better) for real datasets, for a $100$-NN procedure predicting low-dimensional categorical $\mathbf{y}$ from LV $\mathbf{\hat{U}}_{te}$. For the Cars dataset $\mathbf{y}$ is a car’s condition (excellent, good, NA, other). For the News dataset $\mathbf{y}$ is a news item’s topic in (Microsoft, Palestine, Obama, economics). For the Rossmann dataset $\mathbf{y}$ is the store's type (a, b, c, d). The test AUC when using {\em all features} is reported as well, bold results are non-inferior to the best result in a paired t-test.}
  \label{tab:downstream_cars}
  \begin{adjustbox}{width=\columnwidth,center}
  \centering
  \begin{tabular}{l|l|l|llll|l}
			\hline
			Dataset & type & d & PCA-Ig. & VAE-Ig. & VAE-Em. & LMMVAE & ~ \\ \hline
		Cars & \makecell{spatial + \\ categorical} & 1 & 0.790 (.001) & 0.789 (.001) & 0.797 (.001) & \textbf{0.849 (.001)} & ~ \\
			& & 2 & 0.843 (.000) & 0.840 (.001) & 0.850 (.001) & \textbf{0.865 (.001)} & ~ \\
			& & 5 & \textbf{0.875 (.000)} & \textbf{0.874 (.001)} & \textbf{0.877 (.001)} & \textbf{0.873 (.002)} & ~ \\
   \cline{3-8}
			& & \textbf{all} & ~ & ~ & ~ & ~ & 0.750 \\ \hline
   News & categorical & 1 & 0.892 (.000) & 0.960 (.005) & 0.959 (.002) & \textbf{0.976 (.002)} & ~ \\
			& & 2 & 0.951 (.000) & 0.963 (.001) & 0.961 (.001) & \textbf{0.975 (.001)} & ~ \\
			& & 5 & \textbf{0.957 (.000)} & 0.946 (.001) & 0.945 (.001) & \textbf{0.962 (.002)} & ~ \\
   \cline{3-8}
			& & \textbf{all} & ~ & ~ & ~ & ~ & 0.729 \\ \hline
   Rossmann & longitudinal & 1 & 0.807 (.002) & 0.799 (.001) & 0.820 (.003) & \textbf{0.827 (.001)} & ~ \\
			& & 2 & 0.811 (.001) & \textbf{0.830 (.001)} & \textbf{0.834 (.001)} & \textbf{0.832 (.001)} & ~ \\
			& & 5 & 0.813 (.001) & 0.836 (.001) & \textbf{0.846 (.002)} & \textbf{0.845 (.001)} & ~ \\
   \cline{3-8}
			& & \textbf{all} & ~ & ~ & ~ & ~ & 0.912 \\ \hline
\end{tabular}
  \end{adjustbox}
\end{table}

\section{Conclusion}
\label{sec:conclusion}
In this study we proposed a novel model and architecture for dimensionality reduction of large datasets, which often violate the assumption of independence between rows. LMMVAE modifies the classic VAE and allows the practitioner to specify the correlation pattern in the data with a much more expressive latent model through separating the latent space into fixed and random parts, and through the use of RE with varied priors. This in turn leads to lower reconstruction error on unseen data in comparison to the latest SOTA methods, and ultimately to more meaningful representations, as was demonstrated in extensive simulations and with real large datasets. Three common covariance structures were discussed and tested: high-cardinality categorical features, longitudinal and spatial features, as well as a combination of those. In the future we hope to explore the usefulness of the resulting learned LV in additional downstream tasks such as visualization, clustering and feature selection.

\begin{ack}
This study was supported in part by a fellowship from the Edmond J. Safra Center for Bioinformatics at Tel-Aviv University, by Israel Science Foundation grant 2180/20 and by Israel Council for Higher Education Data-Science Centers. UK Biobank research has been conducted using the UK Biobank Resource under Application Number 56885.
\end{ack}

\newpage
\bibliographystyle{plainnat}
\bibliography{references}

\begin{thebibliography}{43}
\providecommand{\natexlab}[1]{#1}
\providecommand{\url}[1]{\texttt{#1}}
\expandafter\ifx\csname urlstyle\endcsname\relax
  \providecommand{\doi}[1]{doi: #1}\else
  \providecommand{\doi}{doi: \begingroup \urlstyle{rm}\Url}\fi

\bibitem[Abadi et~al.(2015)Abadi, Agarwal, Barham, Brevdo, Chen, Citro, Corrado, Davis, Dean, Devin, Ghemawat, Goodfellow, Harp, Irving, Isard, Jia, Jozefowicz, Kaiser, Kudlur, Levenberg, Man\'{e}, Monga, Moore, Murray, Olah, Schuster, Shlens, Steiner, Sutskever, Talwar, Tucker, Vanhoucke, Vasudevan, Vi\'{e}gas, Vinyals, Warden, Wattenberg, Wicke, Yu, and Zheng]{tensorflow2015-whitepaper}
Mart\'{\i}n Abadi, Ashish Agarwal, Paul Barham, Eugene Brevdo, Zhifeng Chen, Craig Citro, Greg~S. Corrado, Andy Davis, Jeffrey Dean, Matthieu Devin, Sanjay Ghemawat, Ian Goodfellow, Andrew Harp, Geoffrey Irving, Michael Isard, Yangqing Jia, Rafal Jozefowicz, Lukasz Kaiser, Manjunath Kudlur, Josh Levenberg, Dandelion Man\'{e}, Rajat Monga, Sherry Moore, Derek Murray, Chris Olah, Mike Schuster, Jonathon Shlens, Benoit Steiner, Ilya Sutskever, Kunal Talwar, Paul Tucker, Vincent Vanhoucke, Vijay Vasudevan, Fernanda Vi\'{e}gas, Oriol Vinyals, Pete Warden, Martin Wattenberg, Martin Wicke, Yuan Yu, and Xiaoqiang Zheng.
\newblock {TensorFlow}: Large-scale machine learning on heterogeneous systems, 2015.
\newblock URL \url{https://www.tensorflow.org/}.
\newblock Software available from tensorflow.org.

\bibitem[Bergeron et~al.(2022)Bergeron, Fung, Hull, Poulos, and Veneris]{Bergeron125}
Maxime Bergeron, Nicholas Fung, John Hull, Zissis Poulos, and Andreas Veneris.
\newblock Variational autoencoders: A hands-off approach to volatility.
\newblock \emph{The Journal of Financial Data Science}, 4\penalty0 (2):\penalty0 125--138, 2022.
\newblock ISSN 2640-3943.
\newblock \doi{10.3905/jfds.2022.1.093}.
\newblock URL \url{https://jfds.pm-research.com/content/4/2/125}.

\bibitem[Bowman et~al.(2015)Bowman, Vilnis, Vinyals, Dai, Jozefowicz, and Bengio]{Bowman2015}
Samuel~R. Bowman, Luke Vilnis, Oriol Vinyals, Andrew~M. Dai, Rafal Jozefowicz, and Samy Bengio.
\newblock Generating sentences from a continuous space, 2015.
\newblock URL \url{https://arxiv.org/abs/1511.06349}.

\bibitem[Casale et~al.(2018)Casale, Dalca, Saglietti, Listgarten, and Fusi]{GPPVAE}
Francesco~Paolo Casale, Adrian Dalca, Luca Saglietti, Jennifer Listgarten, and Nicolo Fusi.
\newblock Gaussian process prior variational autoencoders.
\newblock In S.~Bengio, H.~Wallach, H.~Larochelle, K.~Grauman, N.~Cesa-Bianchi, and R.~Garnett, editors, \emph{Advances in Neural Information Processing Systems}, volume~31. Curran Associates, Inc., 2018.
\newblock URL \url{https://proceedings.neurips.cc/paper/2018/file/1c336b8080f82bcc2cd2499b4c57261d-Paper.pdf}.

\bibitem[CDC(2017)]{asthma}
CDC.
\newblock National environmental public health tracking network data explorer - asthma in adults, Nov 2017.
\newblock URL \url{https://www.cdc.gov/nceh/tracking/topics/asthma.htm}.

\bibitem[Chollet et~al.(2015)]{chollet2015keras}
Fran\c{c}ois Chollet et~al.
\newblock Keras.
\newblock \url{https://keras.io}, 2015.

\bibitem[Cressie(1993)]{Cressie1993}
Noel A.~C. Cressie.
\newblock \emph{Statistics for spatial data}.
\newblock Wiley series in probability and statistics. Wiley-Interscience Publication, New York, revised edition.. edition, 1993.
\newblock ISBN 1-119-11515-9.

\bibitem[Curi et~al.(2019)Curi, Converse, Hajewski, and Oliveira]{Curi2019}
Mariana Curi, Geoffrey~A. Converse, Jeff Hajewski, and Suely Oliveira.
\newblock Interpretable variational autoencoders for cognitive models.
\newblock In \emph{2019 International Joint Conference on Neural Networks (IJCNN)}, pages 1--8, 2019.
\newblock \doi{10.1109/IJCNN.2019.8852333}.

\bibitem[Doersch(2016)]{Doersch16}
Carl Doersch.
\newblock Tutorial on variational autoencoders, 2016.
\newblock URL \url{https://arxiv.org/abs/1606.05908}.

\bibitem[Fabius and Van~Amersfoort(2014)]{VRAE}
Otto Fabius and Joost~R Van~Amersfoort.
\newblock Variational recurrent auto-encoders.
\newblock \emph{arXiv preprint arXiv:1412.6581}, 2014.

\bibitem[Gupta and Nagar(2018)]{gupta2018matrix}
Arjun~K Gupta and Daya~K Nagar.
\newblock \emph{Matrix variate distributions}.
\newblock Chapman and Hall/CRC, 2018.

\bibitem[Hancock and Khoshgoftaar(2020)]{Hancock2020}
John~T. Hancock and Taghi~M. Khoshgoftaar.
\newblock Survey on categorical data for neural networks.
\newblock \emph{Journal of Big Data}, 7\penalty0 (1):\penalty0 28, Apr 2020.
\newblock ISSN 2196-1115.
\newblock \doi{10.1186/s40537-020-00305-w}.
\newblock URL \url{https://doi.org/10.1186/s40537-020-00305-w}.

\bibitem[Higgins et~al.(2017)Higgins, Matthey, Pal, Burgess, Glorot, Botvinick, Mohamed, and Lerchner]{betaVAE}
Irina Higgins, Loic Matthey, Arka Pal, Christopher Burgess, Xavier Glorot, Matthew Botvinick, Shakir Mohamed, and Alexander Lerchner.
\newblock beta-{VAE}: Learning basic visual concepts with a constrained variational framework.
\newblock In \emph{International Conference on Learning Representations}, 2017.
\newblock URL \url{https://openreview.net/forum?id=Sy2fzU9gl}.

\bibitem[Hinton and Salakhutdinov(2006)]{hinton2006}
Geoffrey~E Hinton and Ruslan~R Salakhutdinov.
\newblock Reducing the dimensionality of data with neural networks.
\newblock \emph{science}, 313\penalty0 (5786):\penalty0 504--507, 2006.

\bibitem[Jazbec et~al.(2021)Jazbec, Ashman, Fortuin, Pearce, Mandt, and R{\"a}tsch]{SVGPVAE}
Metod Jazbec, Matt Ashman, Vincent Fortuin, Michael Pearce, Stephan Mandt, and Gunnar R{\"a}tsch.
\newblock Scalable gaussian process variational autoencoders.
\newblock In Arindam Banerjee and Kenji Fukumizu, editors, \emph{Proceedings of The 24th International Conference on Artificial Intelligence and Statistics}, volume 130 of \emph{Proceedings of Machine Learning Research}, pages 3511--3519. PMLR, 13--15 Apr 2021.
\newblock URL \url{https://proceedings.mlr.press/v130/jazbec21a.html}.

\bibitem[Jolliffe(2002)]{jolliffe2002}
I.T. Jolliffe.
\newblock \emph{Principal Component Analysis}.
\newblock Springer Series in Statistics. Springer, 2002.
\newblock ISBN 9780387954424.

\bibitem[Kalehbasti et~al.(2019)Kalehbasti, Nikolenko, and Rezaei]{kalehbasti2019airbnb}
Pouya~Rezazadeh Kalehbasti, Liubov Nikolenko, and Hoormazd Rezaei.
\newblock Airbnb price prediction using machine learning and sentiment analysis, 2019.

\bibitem[Kingma and Welling(2013)]{VAE2013}
Diederik~P Kingma and Max Welling.
\newblock Auto-encoding variational bayes, 2013.
\newblock URL \url{https://arxiv.org/abs/1312.6114}.

\bibitem[Kingma and Welling(2019)]{VAE2019}
Diederik~P. Kingma and Max Welling.
\newblock An introduction to variational autoencoders.
\newblock \emph{Foundations and Trends® in Machine Learning}, 12\penalty0 (4):\penalty0 307--392, 2019.
\newblock ISSN 1935-8237.
\newblock \doi{10.1561/2200000056}.
\newblock URL \url{http://dx.doi.org/10.1561/2200000056}.

\bibitem[Lawrence(2005)]{Lawrence05}
Neil Lawrence.
\newblock Probabilistic non-linear principal component analysis with gaussian process latent variable models.
\newblock \emph{Journal of Machine Learning Research}, 6\penalty0 (60):\penalty0 1783--1816, 2005.
\newblock URL \url{http://jmlr.org/papers/v6/lawrence05a.html}.

\bibitem[Liu et~al.(2015)Liu, Luo, Wang, and Tang]{liu2015faceattributes}
Ziwei Liu, Ping Luo, Xiaogang Wang, and Xiaoou Tang.
\newblock Deep learning face attributes in the wild.
\newblock In \emph{Proceedings of International Conference on Computer Vision (ICCV)}, December 2015.

\bibitem[Maaløe et~al.(2016)Maaløe, Sønderby, Sønderby, and Winther]{ADGM}
Lars Maaløe, Casper~Kaae Sønderby, Søren~Kaae Sønderby, and Ole Winther.
\newblock Auxiliary deep generative models.
\newblock In Maria~Florina Balcan and Kilian~Q. Weinberger, editors, \emph{Proceedings of The 33rd International Conference on Machine Learning}, volume~48 of \emph{Proceedings of Machine Learning Research}, pages 1445--1453, New York, New York, USA, 20--22 Jun 2016. PMLR.
\newblock URL \url{https://proceedings.mlr.press/v48/maaloe16.html}.

\bibitem[McCulloch et~al.(2008)McCulloch, Searle, and Neuhaus]{McCulloch2008}
Charles~E. McCulloch, Shayle~R. Searle, and John~M. Neuhaus.
\newblock \emph{Generalized, Linear, and Mixed Models}.
\newblock John Wiley and Sons, Inc., June 2008.
\newblock ISBN 978-0-470-07371-1.

\bibitem[Mock(2022)]{tidytuesday}
Thomas Mock.
\newblock Tidy tuesday: A weekly data project aimed at the r ecosystem, 2022.
\newblock URL \url{https://github.com/rfordatascience/tidytuesday}.

\bibitem[Moniz and Torgo(2018)]{Moniz2018}
Nuno Moniz and Luis Torgo.
\newblock Multi-source social feedback of online news feeds.
\newblock \emph{CoRR}, 2018.

\bibitem[MuonNeutrino(2019)]{us_census}
MuonNeutrino.
\newblock Us census demographic data, Mar 2019.
\newblock URL \url{https://www.kaggle.com/datasets/muonneutrino/us-census-demographic-data}.

\bibitem[Qiang et~al.(2020)Qiang, Dong, Sun, Ge, and Liu]{Ning2020}
Ning Qiang, Qinglin Dong, Yifei Sun, Bao Ge, and Tianming Liu.
\newblock Deep variational autoencoder for modeling functional brain networks and adhd identification.
\newblock In \emph{2020 IEEE 17th International Symposium on Biomedical Imaging (ISBI)}, pages 554--557, 2020.
\newblock \doi{10.1109/ISBI45749.2020.9098480}.

\bibitem[Qui{{\~n}}onero-Candela and Rasmussen(2005)]{JMLR:v6:quinonero-candela05a}
Joaquin Qui{{\~n}}onero-Candela and Carl~Edward Rasmussen.
\newblock A unifying view of sparse approximate gaussian process regression.
\newblock \emph{Journal of Machine Learning Research}, 6\penalty0 (65):\penalty0 1939--1959, 2005.
\newblock URL \url{http://jmlr.org/papers/v6/quinonero-candela05a.html}.

\bibitem[Ramsay and Silverman(2005)]{ramsay2005functional}
J.~O. Ramsay and B.W. Silverman.
\newblock \emph{Functional Data Analysis}.
\newblock Springer Series in Statistics. Springer, 2005.
\newblock ISBN 9780387400808.

\bibitem[Ranganath et~al.(2016)Ranganath, Tran, and Blei]{HVM}
Rajesh Ranganath, Dustin Tran, and David Blei.
\newblock Hierarchical variational models.
\newblock In Maria~Florina Balcan and Kilian~Q. Weinberger, editors, \emph{Proceedings of The 33rd International Conference on Machine Learning}, volume~48 of \emph{Proceedings of Machine Learning Research}, pages 324--333, New York, New York, USA, 20--22 Jun 2016. PMLR.
\newblock URL \url{https://proceedings.mlr.press/v48/ranganath16.html}.

\bibitem[Rasmussen and Williams(2005)]{Rasmussen}
Carl~Edward Rasmussen and Christopher K.~I. Williams.
\newblock \emph{Gaussian Processes for Machine Learning (Adaptive Computation and Machine Learning)}.
\newblock The MIT Press, 2005.
\newblock ISBN 026218253X.

\bibitem[Reese(2020)]{Cars}
Austin Reese.
\newblock Used cars dataset - vehicles listings from craigslist.org, 2020.
\newblock URL \url{https://www.kaggle.com/datasets/austinreese/craigslist-carstrucks-data}.

\bibitem[Rolinek et~al.(2018)Rolinek, Zietlow, and Martius]{Rolinek2018}
Michal Rolinek, Dominik Zietlow, and Georg Martius.
\newblock Variational autoencoders pursue pca directions (by accident), 2018.
\newblock URL \url{https://arxiv.org/abs/1812.06775}.

\bibitem[Rossmann(2016)]{rossmann}
Rossmann.
\newblock Rossmann store sales, 2016.
\newblock URL \url{https://www.kaggle.com/competitions/rossmann-store-sales/}.

\bibitem[Sch{\"o}lkopf et~al.(1998)Sch{\"o}lkopf, Smola, and M{\"u}ller]{KPCA}
Bernhard Sch{\"o}lkopf, Alexander Smola, and Klaus-Robert M{\"u}ller.
\newblock Nonlinear component analysis as a kernel eigenvalue problem.
\newblock \emph{Neural computation}, 10\penalty0 (5):\penalty0 1299--1319, 1998.

\bibitem[Searle et~al.(1992)Searle, Casella, and McCulloch]{searle1992variance}
Shayle~R Searle, George Casella, and Charles McCulloch.
\newblock \emph{Variance components}.
\newblock Wiley Series in Probability and Statistics. John Wiley \& Sons, 1992.

\bibitem[Simchoni and Rosset(2021)]{lmmnn_neurips}
Giora Simchoni and Saharon Rosset.
\newblock Using random effects to account for high-cardinality categorical features and repeated measures in deep neural networks.
\newblock In M.~Ranzato, A.~Beygelzimer, Y.~Dauphin, P.S. Liang, and J.~Wortman Vaughan, editors, \emph{Advances in Neural Information Processing Systems}, volume~34, pages 25111--25122. Curran Associates, Inc., 2021.
\newblock URL \url{https://proceedings.neurips.cc/paper/2021/file/d35b05a832e2bb91f110d54e34e2da79-Paper.pdf}.

\bibitem[Simchoni and Rosset(2023)]{LMMNN}
Giora Simchoni and Saharon Rosset.
\newblock Integrating random effects in deep neural networks.
\newblock \emph{Journal of Machine Learning Research}, 2023.
\newblock \doi{10.48550/ARXIV.2206.03314}.
\newblock URL \url{https://arxiv.org/abs/2206.03314}.
\newblock to appear.

\bibitem[Sudlow et~al.(2015)Sudlow, Gallacher, Allen, Beral, Burton, Danesh, Downey, Elliott, Green, Landray, Liu, Matthews, Ong, Pell, Silman, Young, Sprosen, Peakman, and Collins]{UKB}
Cathie Sudlow, John Gallacher, Naomi Allen, Valerie Beral, Paul Burton, John Danesh, Paul Downey, Paul Elliott, Jane Green, Martin Landray, Bette Liu, Paul Matthews, Giok Ong, Jill Pell, Alan Silman, Alan Young, Tim Sprosen, Tim Peakman, and Rory Collins.
\newblock Uk biobank: An open access resource for identifying the causes of a wide range of complex diseases of middle and old age.
\newblock \emph{PLOS Medicine}, 12\penalty0 (3):\penalty0 1--10, 03 2015.
\newblock \doi{10.1371/journal.pmed.1001779}.
\newblock URL \url{https://doi.org/10.1371/journal.pmed.1001779}.

\bibitem[Tipping and Bishop(1999)]{PPCA}
Michael~E. Tipping and Christopher~M. Bishop.
\newblock Probabilistic principal component analysis.
\newblock \emph{Journal of the Royal Statistical Society. Series B (Statistical Methodology)}, 61\penalty0 (3):\penalty0 611--622, 1999.
\newblock ISSN 13697412, 14679868.
\newblock URL \url{http://www.jstor.org/stable/2680726}.

\bibitem[Tran et~al.(2020)Tran, Nguyen, Nott, and Kohn]{DeepGLMM}
Minh-Ngoc Tran, Nghia Nguyen, David Nott, and Robert Kohn.
\newblock Bayesian deep net glm and glmm.
\newblock \emph{Journal of Computational and Graphical Statistics}, 29\penalty0 (1):\penalty0 97--113, 2020.
\newblock \doi{10.1080/10618600.2019.1637747}.
\newblock URL \url{https://doi.org/10.1080/10618600.2019.1637747}.

\bibitem[Wei and Mahmood(2021)]{Wei2021Review}
Ruoqi Wei and Ausif Mahmood.
\newblock Recent advances in variational autoencoders with representation learning for biomedical informatics: A survey.
\newblock \emph{IEEE Access}, 9:\penalty0 4939--4956, 2021.
\newblock \doi{10.1109/ACCESS.2020.3048309}.

\bibitem[Xiong et~al.(2019)Xiong, Kim, and Singh]{MeNets}
Yunyang Xiong, Hyunwoo~J. Kim, and Vikas Singh.
\newblock Mixed effects neural networks (menets) with applications to gaze estimation.
\newblock In \emph{Proceedings of the IEEE/CVF Conference on Computer Vision and Pattern Recognition (CVPR)}, June 2019.

\end{thebibliography}

\newpage
\appendix
\section{Proof of KL-divergence decomposition}
\label{app:Dkl_decomposition}
Mark $\mathbf{z} = (\mathbf{u}, \mathbf{b})$, the entire set of LV for LMMVAE. Then assuming $\mathbf{u}$ and $\mathbf{b}$ to be both marginally independent and conditionally independent given the data, we utilize separate encoders in \eqref{eq:independece}:
\begin{align*}
    D_{KL}[q(\mathbf{z}|\mathbf{x})||p(\mathbf{z})] &= D_{KL}[q(\mathbf{u}|\mathbf{x})q(\mathbf{b}|\mathbf{x})||p(\mathbf{u})p(\mathbf{b})] \\
		&= \int \left[q(\mathbf{u}|\mathbf{x})q(\mathbf{b}|\mathbf{x})\log\frac{q(\mathbf{u}|\mathbf{x})q(\mathbf{b}|\mathbf{x})}{p(\mathbf{u})p(\mathbf{b})}\right]d{u}d{b} \\
		&= \int\left[q(\mathbf{u}|\mathbf{x})q(\mathbf{b}|\mathbf{x})\log\frac{q(\mathbf{u}|\mathbf{x})}{p(\mathbf{u})}\right]d{u}d{b} \\ & + \int\left[q(\mathbf{u}|\mathbf{x})q(\mathbf{b}|\mathbf{x})\log\frac{q(\mathbf{b}|\mathbf{x})}{p(\mathbf{b})}\right]d{u}d{b} \\
		&= \int q(\mathbf{b}|\mathbf{x})d{b} \int q(\mathbf{u}|\mathbf{x})\log\frac{q(\mathbf{u}|\mathbf{x})}{p(\mathbf{u})}d{u} \\
            & + \int q(\mathbf{u}|\mathbf{x})d{u} \int q(\mathbf{b}|\mathbf{x})\log\frac{q(\mathbf{b}|\mathbf{x})}{p(\mathbf{b})}d{b} \\
		&= D_{KL}[q(\mathbf{u}|\mathbf{x})||p(\mathbf{u})] + D_{KL}[q(\mathbf{b}|\mathbf{x})||p(\mathbf{b})]
\end{align*}

Hence, if we write the general ELBO as in \eqref{eq:ELBO1}, replace $\mathbf{u}$ with a general $\mathbf{z}$ and replace $D_{KL}[q(\mathbf{z}|\mathbf{x})||p(\mathbf{z})]$ with the above decomposition, we get \eqref{eq:ELBO2}.

\section{Deriving LMMVAE Loss}
\label{app:Loss}
The development of the first two terms in \eqref{eq:Loss0} is a known result \citep[see e.g.][]{Doersch16}. We are left with developing the third term, which is the KL-divergence between RE $\mathbf{b}$'s prior $p(\mathbf{b})$ and the surrogate posterior $q_\phi(\mathbf{b}|\mathbf{x})$, in the single categorical case. As described in Section~\ref{sec:LMMVAE} $p(\mathbf{b}_j)$ for cluster $j$'s observations is $\mathcal{N}(\mathbf{0}, \mathbf{D})$, where $\mathbf{D} = \text{diag}(\sigma^2_{b1}, \dots, \sigma^2_{bp})$. We use a standard diagonal Gaussian $q_\phi(\mathbf{b}_j|\mathbf{x})$ surrogate posterior $\mathcal{N}(\bm{\mu}_b, \text{diag}(\bm{\tau}^2_b))$, where $\bm{\mu}_b \in \mathbb{R}^p$ and $\bm{\tau}^2_b \in \mathbb{R}^p$ are output by the DNN encoder. The KL-divergence then is between two Gaussian distributions:

\begin{align*}
    & D_{KL}[q(\mathbf{b}|\mathbf{x})||p(\mathbf{b})] = \\ &= \frac{1}{2}\left[\log\frac{|\mathbf{D}|}{|\bm{\Sigma}|} - p + \text{Tr} \{ \mathbf{D}^{-1}\bm{\Sigma} \} + (\mathbf{0} - \bm{\mu}_b)^T \mathbf{D}^{-1}(\mathbf{0} - \bm{\mu}_b)\right] \\
    &= \frac{1}{2}\left[\log{|\mathbf{D}|}-\log{|\bm{\Sigma}|} - p + \text{Tr} \{ \mathbf{D}^{-1}\bm{\Sigma} \} + \bm{\mu}_b^T\mathbf{D}^{-1} \bm{\mu}_b\right] \\
    &= \frac{1}{2}\left[\sum_{k=1}^{p}\log\sigma^2_{bk}-\sum_{k=1}^{p}\log\tau^2_{bk} - p + \sum_{k=1}^{p}\frac{\tau^2_{bk}}{\sigma^2_{bk}} + \sum_{k=1}^{p}\frac{\mu_{bk}^2}{\sigma^2_{bk}}\right]\\
    &= -\frac{1}{2}\sum_{k=1}^{p}\left[1 -\log\sigma_{bk}^2 + \log\tau_{bk}^2 - \frac{\tau^2_{bk}}{\sigma^2_{bk}} - \frac{\mu_{bk}^2}{\sigma^2_{bk}}\right]\\
    &= -\frac{1}{2}\sum_{k=1}^{p}\left[1 + \gamma_{bk} -\delta_{bk} - e^{\gamma_{bk} - \delta_{bk}} - \mu^2_{bk}e^{-\delta_{bk}}\right],
\end{align*}
where $\bm{\gamma}_b = \log \bm{\tau}^2_b$, the actual output of the DNN encoder, and  $\bm{\delta}_b = \log \bm{\sigma}^2_b$, the log of the prior RE variance components in $\mathbf{D}$. Multiplying the above by $\beta$ for the $\beta$-VAE version gives the final error term in \eqref{eq:Loss0}.

\newpage
\section{Real datasets additional details}
\label{app:realdetails}
\begin{table}[H]
  \caption{Real datasets: additional details}
  \label{app:tab:real-data-details}
  \centering
  \begin{tabular}
{p{0.1\linewidth}p{0.15\linewidth}p{0.15\linewidth}p{0.5\linewidth}}
    \toprule
    \multicolumn{4}{c}{\textbf{Categorical}} \\
    \midrule
    Dataset & Source & Reference & Description \\
    \midrule
    News & UCI ML (Free) & \cite{Moniz2018} & 81K news items and their no. of shares on Facebook. Features are top 1-gram tokens count in headline. High-cardinality categorical features include source ($q = 5.4K$) and unique title ($q = 72K$).\\
    Spotify & Tidy Tuesday (Free) & \cite{tidytuesday} & 28K songs with 12 audio features. High-cardinality categorical features include artist ($q = 10K$), album ($q = 22K$), playlist ($q = 2.3K$) and subgenre ($q = 553$).\\
    \midrule
    \multicolumn{4}{c}{\textbf{Longitudinal}} \\
    \midrule
    Rossmann & Kaggle (Free) & \cite{rossmann} & Total monthly sales in \$ from over $q = 1.1K$ stores around Europe, over 25-31 months. Features include month, number of holiday days, number of days with promotion and more. \\
    UKB & UK Biobank (Permission) & \cite{UKB} & $q=469K$ subjects of the UK Biobank cohort for which we have 1-4 systolic blood pressure (SBP) measures. Time-varying features include gender, age, height, different food intakes, smoking habits and many more. \\
    \midrule
    \multicolumn{4}{c}{\textbf{Spatial}} \\
    \midrule
    Income & Kaggle (Free) & \cite{us_census} & Mean yearly income in \$ for 71K US census tracts grouped by $q = 3K$ locations, data was previously downloaded from the US Census Bureau. In addition to longitude and latitude features include population size, rate of employment and more. \\
    Asthma & CDC (Free) & \cite{asthma} & Adult asthma rate in 69K US census tracts according to CDC in 2019, grouped by $q = 3K$ locations. Additional features come from the income data. \\
    \midrule
    \multicolumn{4}{c}{\textbf{Spatial-Categorical combination}} \\
    \midrule
    Cars & Kaggle (Free) & \cite{Cars} & 97K cars with unique VIN from Craigslist and their price in \$, price was filtered from 1K\$ to 300K\$. In addition to longitude and latitude (overall $q = 12K$ unique locations), features include manufacturer, year of make, size and more, as well as the car's model ($q = 15K$). \\
    Airbnb & Google Drive (Free) & \cite{kalehbasti2019airbnb} & 50K Airbnb listings in NYC scraped by \cite{kalehbasti2019airbnb}, ETL follows their steps exactly. In addition to longitude and latitude (overall $q = 3K$ unique locations), features include floor number, neighborhood, some top 1-ngram tokens counts from description and more, as well as the listing's host ($q = 40K$). \\
    \midrule
    \multicolumn{4}{c}{\textbf{Image data}} \\
    \midrule
    CelebA & Google Drive (Free) & \cite{liu2015faceattributes} & 202K facial images from 10,177 celebrities ($q = 10K$). Additional features not considered here are binary features such as ``wears glasses'' and continuous features such as pixel nose locations. \\
    \bottomrule
 \end{tabular}
\end{table}

\newpage
\section{Real Data: Additional Visualizations}
\label{app:additional:real:viz}

\begin{figure}[H]
    \centering
    \includegraphics[width=1.0\linewidth]{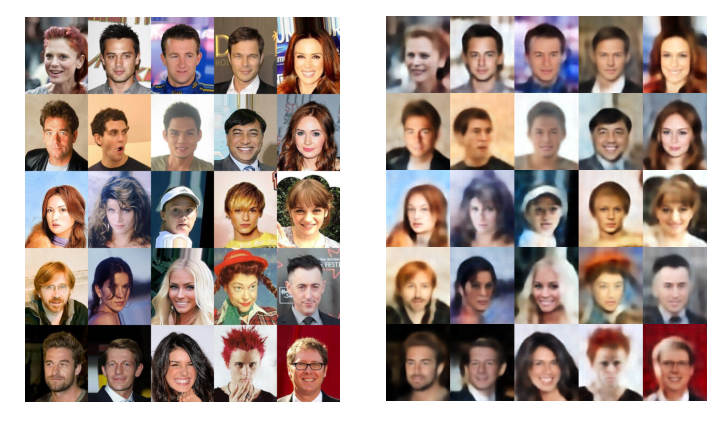}
    \caption{Comparing true vs. reconstructed $\mathbf{X}_{te}$ for the CelebA dataset of facial images with $d = 100$, using convolutional neural networks for both FE and RE encoders. Left: true faces; Right: reconstructed faces.}
    \label{app:fig:additional:real:Celeba_recon}
\end{figure}

\newpage
\section{Simulated Data: Additional Results}
\label{app:additional}

\begin{table}[H]
  \caption{Simulated datasets with high-cardinality categorical features: mean test reconstruction errors for $n = 100000$ observations, $p = 100$ fixed features, 3 categorical features, with $q_1 = 1000, q_2 = 3000, q_3 = 5000$, standard errors in parentheses. Bold results are non-inferior to the best result in a paired t-test.}
  \label{app:tab:additional:simulated:categorical:mse}
  \begin{adjustbox}{width=\columnwidth,center}
  \centering
  \begin{tabular}{lll|llllll}
\toprule
$\sigma^2_{b_1}$ & $\sigma^2_{b_2}$ & $\sigma^2_{b_3}$ & PCA-Ignore & PCA-OHE & VAE-Ignore & VAE-OHE & VAE-Embed. & LMMVAE \\
  \midrule
  \multicolumn{9}{c}{\textbf{d = 1}} \\
	\midrule
  0.3 & 0.3 & 0.3 & 2.31 (.02) & 2.31 (.02) & 2.15 (.01) & 2.24 (.01) & 2.16 (.01) & \textbf{1.28 (.02)} \\
  & & 3.0 & 5.86 (.10) & 5.90 (.09) & 5.70 (.10) & 5.72 (.05) & 5.69 (.10) & \textbf{1.37 (.01)} \\
  & 3.0 & 0.3 & 5.96 (.04) & 6.00 (.03) & 5.76 (.04) & 5.83 (.08) & 5.78 (.04) & \textbf{1.52 (.02)} \\
  & & 3.0 & 9.50 (.11) & 9.61 (.12) & 9.24 (.09) & 9.33 (.11) & 9.28 (.09) & \textbf{2.01 (.05)} \\
	\midrule
  3.0 & 0.3 & 0.3 & 5.95 (.05) & 6.01 (.04) & 5.78 (.05) & 5.84 (.16) & 5.75 (.05) & \textbf{1.57 (.01)} \\
  & & 3.0 & 9.50 (.15) & 9.64 (.14) & 9.23 (.13) & 9.32 (.08) & 9.25 (.14) & \textbf{1.70 (.03)} \\
  & 3.0 & 0.3 & 9.51 (.07) & 9.63 (.08) & 9.24 (.08) & 9.37 (.08) & 9.26 (.07) & \textbf{1.89 (.03)} \\
  & & 3.0 & 13.0 (.08) & 13.1 (.08) & 12.7 (.08) & 12.9 (.17) & 12.6 (.08) & \textbf{2.23 (.01)} \\
	\midrule
	\multicolumn{9}{c}{\textbf{d = 2}} \\
	\midrule
  0.3 & 0.3 & 0.3 & 2.56 (.02) & 2.58 (.02) & 2.21 (.03) & 2.32 (.02) & 2.26 (.03) & \textbf{1.52 (.04)} \\
  & & 3.0 & 5.95 (.08) & 6.04 (.08) & 5.65 (.08) & 5.86 (.03) & 5.69 (.09) & \textbf{1.59 (.04)} \\
  & 3.0 & 0.3 & 5.92 (.19) & 6.02 (.20) & 5.60 (.18) & 5.79 (.05) & 5.65 (.19) & \textbf{1.83 (.05)} \\
  & & 3.0 & 9.75 (.15) & 10.0 (.13) & 9.26 (.14) & 9.09 (.03) & 9.34 (.13) & \textbf{2.28 (.06)} \\
	\midrule
  3.0 & 0.3 & 0.3 & 6.07 (.07) & 6.17 (.08) & 5.77 (.10) & 5.68 (.05) & 5.72 (.08) & \textbf{1.81 (.07)} \\
  & & 3.0 & 9.49 (.06) & 9.73 (.05) & 9.05 (.08) & 9.40 (.11) & 9.20 (.07) & \textbf{1.93 (.03)} \\
  & 3.0 & 0.3 & 9.47 (.11) & 9.72 (.11) & 8.99 (.08) & 9.20 (.13) & 9.03 (.12) & \textbf{2.05 (.02)} \\
  & & 3.0 & 13.1 (.15) & 13.5 (.16) & 12.6 (.20) & 12.5 (.15) & 12.7 (.08) & \textbf{2.49 (.06)} \\
\bottomrule
\end{tabular}
  \end{adjustbox}
\end{table}

\begin{table}[H]
    \caption{Simulated datasets with high-cardinality categorical features: Mean running times (seconds) for $n = 100000$ observations, $p = 100$ fixed features, 3 categorical features, with $q_1 = 1000, q_2 = 3000, q_3 = 5000$. All DNN experiments ran with 200 epochs.}
    \label{app:tab:additional:simulated:categorical:time}
    \centering
    \begin{tabular}{lll|llllll}
\toprule
$\sigma^2_{b_1}$ & $\sigma^2_{b_2}$ & $\sigma^2_{b_3}$ & PCA-Ignore & PCA-OHE & VAE-Ignore & VAE-OHE & VAE-Embed. & LMMVAE \\
    \midrule
    \multicolumn{9}{c}{\textbf{d = 1}} \\
	\midrule
    0.3 & 0.3 & 0.3 & 1.0 & 216.2 & 93.9 & 456.0 & 140.0 & 226.7 \\
    &  & 3.0 & 1.0 & 226.2 & 91.9 & 450.2 & 141.1 & 226.5 \\
    & 3.0 & 0.3 & 1.0 & 223.9 & 92.7 & 459.7 & 141.4 & 227.4 \\
    &  & 3.0 & 1.2 & 226.1 & 96.4 & 456.3 & 145.1 & 243.7 \\
	\midrule
    3.0 & 0.3 & 0.3 & 1.0 & 227.7 & 91.5 & 450.6 & 144.3 & 225.1 \\
    &  & 3.0 & 1.1 & 227.7 & 93.1 & 454.6 & 142.0 & 227.4 \\
    & 3.0 & 0.3 & 1.1 & 229.3 & 91.2 & 463.3 & 141.4 & 224.9 \\
    &  & 3.0 & 1.1 & 224.0 & 93.0 & 455.0 & 141.4 & 227.8 \\
	\midrule
	\multicolumn{9}{c}{\textbf{d = 2}} \\
	\midrule
    0.3 & 0.3 & 0.3 & 1.0 & 226.2 & 93.5 & 454.6 & 141.0 & 227.2 \\
    &  & 3.0 & 1.2 & 229.9 & 91.7 & 453.2 & 141.9 & 237.4 \\
    & 3.0 & 0.3 & 1.1 & 227.1 & 95.1 & 452.7 & 144.0 & 228.9 \\
    &  & 3.0 & 1.1 & 228.3 & 94.3 & 452.8 & 140.5 & 228.7 \\
	\midrule
    3.0 & 0.3 & 1.1 & 225.6 & 96.4 & 457.4 & 144.1 & 229.1 \\
    &  & 3.0 & 1.1 & 229.4 & 95.7 & 460.9 & 145.4 & 229.9 \\
    & 3.0 & 1.1 & 227.5 & 92.5 & 457.9 & 142.6 & 224.7 \\
    &  & 3.0 & 1.1 & 226.9 & 94.9 & 460.7 & 144.0 & 229.6 \\
\bottomrule
\end{tabular}
\end{table}

\begin{table}[H]
    \caption{Simulated datasets with high-cardinality categorical features: mean test NLL loss for $n = 100000$ observations, $p = 100$ fixed features, 3 categorical features, with $q_1 = 1000, q_2 = 3000, q_3 = 5000$.}
    \label{app:tab:additional:simulated:categorical:nll}
    \centering
    \begin{tabular}{lll|lll}
\toprule
$\sigma^2_{b_1}$ & $\sigma^2_{b_2}$ & $\sigma^2_{b_3}$ & VAE-Ignore & VAE-Embed. & LMMVAE \\
    \midrule
    \multicolumn{6}{c}{\textbf{d = 1}} \\
	\midrule
    0.3 & 0.3 & 0.3 & 214.6 & 216.5 & \textbf{96.7} \\
    &  & 3.0 & 570.5 & 569.3 & \textbf{105.1} \\
    & 3.0 & 0.3 & 576.5 & 577.7 & \textbf{117.4} \\
    &  & 3.0 & 924.2 & 928.1 & \textbf{153.0} \\
	\midrule
    3.0 & 0.3 & 0.3 & 578.2 & 575.2 & \textbf{119.3} \\
    &  & 3.0 & 923.1 & 925.2 & \textbf{129.3} \\
    & 3.0 & 0.3 & 923.6 & 926.4 & \textbf{144.6} \\
    &  & 3.0 & 1268.2 & 1260.8 & \textbf{169.2} \\
	\midrule
	\multicolumn{6}{c}{\textbf{d = 2}} \\
	\midrule
    0.3 & 0.3 & 0.3 & 221.5 & 225.8 & \textbf{113.3} \\
    &  & 3.0 & 564.9 & 568.7 & \textbf{120.4} \\
    & 3.0 & 0.3 & 560.2 & 565.0 & \textbf{137.9} \\
    &  & 3.0 & 926.6 & 934.4 & \textbf{171.1} \\
	\midrule
    3.0 & 0.3 & 0.3 & 577.2 & 571.7 & \textbf{136.5} \\
    &  & 3.0 & 905.3 & 920.0 & \textbf{145.7} \\
    & 3.0 & 0.3 & 899.1 & 902.9 & \textbf{155.7} \\
    &  & 3.0 & 1261.5 & 1275.1 & \textbf{187.8} \\
\bottomrule
\end{tabular}
\end{table}

\begin{table}[H]
  \caption{Simulated datasets with longitudinal features: mean test reconstruction errors for $n = 100000$ observations, $p = 100$ fixed features, $q = 1000$ subjects, and $K = 3$ polynomial terms on $t$, random mode. Standard errors in parentheses, bold results are non-inferior to the best result in a paired t-test.}
  \label{app:additional:simulated:longitudinal:mse1000}
  \begin{adjustbox}{width=\columnwidth,center}
  \centering
  \begin{tabular}{lll|lllllll}
\toprule
$\sigma^2_{b_0}$ & $\sigma^2_{b_1}$ & $\sigma^2_{b_2}$ & PCA-Ignore & PCA-OHE & VAE-Ignore & VAE-Embed & SVGPVAE & VRAE & LMMVAE \\
 \midrule
 \multicolumn{10}{c}{\textbf{d = 1}} \\
	\midrule
 0.3 & 0.3 & 0.3 & 1.78 (.02) & 1.82 (.02) & 1.60 (.00) & 1.61 (.00) & 2.10 (.03) & 2.06 (.03) & \textbf{1.11 (.02)} \\
 & & 3.0 & 2.44 (.03) & 2.45 (.03) & 2.29 (.03) & 2.27 (.01) & 2.72 (.03) & 2.68 (.03) & \textbf{1.25 (.01)} \\
 & 3.0 & 0.3 & 2.90 (.02) & 2.90 (.04) & 2.69 (.00) & 2.69 (.01) & 3.07 (.02) & 3.11 (.02) & \textbf{1.29 (.03)} \\
 & & 3.0 & 3.54 (.02) & 3.54 (.03) & 3.38 (.02) & 3.37 (.01) & 3.75 (.03) & 3.85 (.04) & \textbf{1.42 (.04)} \\
	\midrule
 3.0 & 0.3 & 0.3 & 4.78 (.03) & 4.77 (.02) & 4.60 (.01) & 4.60 (.01) & 4.89 (.02) & 5.08 (.02) & \textbf{1.20 (.04)} \\
 & & 3.0 & 5.71 (.03) & 5.74 (.02) & 5.54 (.02) & 5.54 (.01) & 5.82 (.06) & 6.00 (.02) & \textbf{1.26 (.01)} \\
 & 3.0 & 0.3 & 6.27 (.02) & 6.29 (.01) & 6.11 (.01) & 6.13 (.01) & 6.30 (.03) & 6.63 (.05) & \textbf{1.24 (.01)} \\
 & & 3.0 & 7.22 (.02) & 7.22 (.02) & 7.08 (.05) & 7.04 (.02) & 7.22 (.04) & 7.50 (.05) & \textbf{1.40 (.03)} \\
	\midrule
	\multicolumn{10}{c}{\textbf{d = 2}} \\
	\midrule
 0.3 & 0.3 & 0.3 & 1.96 (.03) & 1.98 (.02) & 1.63 (.00) & 1.65 (.02) & 2.46 (.04) & 2.54 (.02) & \textbf{1.11 (.02)} \\
 & & 3.0 & 2.65 (.02) & 2.60 (.01) & 2.30 (.02) & 2.30 (.01) & 3.13 (.04) & 3.17 (.03) & \textbf{1.33 (.04)} \\
 & 3.0 & 0.3 & 3.08 (.02) & 3.03 (.03) & 2.70 (.01) & 2.72 (.02) & 3.43 (.02) & 3.63 (.06) & \textbf{1.25 (.01)} \\
 & & 3.0 & 3.70 (.02) & 3.71 (.04) & 3.35 (.02) & 3.40 (.03) & 4.00 (.07) & 4.23 (.03) & \textbf{1.42 (.01)} \\
	\midrule
 3.0 & 0.3 & 0.3 & 4.93 (.02) & 4.89 (.02) & 4.62 (.03) & 4.65 (.04) & 4.98 (.04) & 5.45 (.07) & \textbf{1.28 (.05)} \\
 & & 3.0 & 5.84 (.02) & 5.87 (.04) & 5.58 (.02) & 5.53 (.02) & 5.77 (.04) & 6.43 (.04) & \textbf{1.56 (.19)} \\
 & 3.0 & 0.3 & 6.37 (.02) & 6.36 (.02) & 6.06 (.05) & 6.09 (.03) & 6.16 (.03) & 7.04 (.04) & \textbf{1.35 (.04)} \\
 & & 3.0 & 7.33 (.02) & 7.36 (.03) & 6.92 (.02) & 6.91 (.05) & 7.00 (.05) & 8.00 (.04) & \textbf{1.47 (.02)} \\
\bottomrule
\end{tabular}
  \end{adjustbox}
\end{table}

\begin{table}[H]
  \caption{Simulated datasets with longitudinal features: mean test reconstruction errors for $n = 100000$ observations, $p = 100$ fixed features, $q = 1000$ subjects, and $K = 3$ polynomial terms on $t$, future mode. Standard errors in parentheses, bold results are non-inferior to the best result in a paired t-test.}
  \label{app:additional:simulated:longitudinal:mse1000:future}
  \begin{adjustbox}{width=\columnwidth,center}
  \centering
  \begin{tabular}{lll|lllllll}
\toprule
$\sigma^2_{b_0}$ & $\sigma^2_{b_1}$ & $\sigma^2_{b_2}$ & PCA-Ignore & PCA-OHE & VAE-Ignore & VAE-Embed & SVGPVAE & VRAE & LMMVAE \\
 \midrule
 \multicolumn{10}{c}{\textbf{d = 1}} \\
	\midrule
 0.3 & 0.3 & 0.3 & 2.26 (.03) & 2.22 (.01) & 2.09 (.03) & 2.07 (.01) & 2.49 (.03) & 2.48 (.01) & \textbf{1.22 (.02)} \\
 & & 3.0 & 4.40 (.02) & 4.38 (.02) & 4.17 (.01) & 4.24 (.03) & 4.61 (.02) & 4.60 (.01) & \textbf{1.99 (.00)} \\
 & 3.0 & 0.3 & 4.74 (.01) & 4.80 (.02) & 4.57 (.01) & 4.62 (.03) & 4.89 (.02) & 5.03 (.04) & \textbf{1.63 (.02)} \\
 & & 3.0 & 6.86 (.02) & 6.88 (.03) & 6.65 (.01) & 6.66 (.01) & 7.03 (.05) & 7.14 (.04) & \textbf{2.27 (.02)} \\
	\midrule
 3.0 & 0.3 & 0.3 & 5.56 (.02) & 5.59 (.03) & 5.38 (.02) & 5.37 (.01) & 5.64 (.05) & 5.83 (.02) & \textbf{1.31 (.08)} \\
 & & 3.0 & 8.35 (.02) & 8.35 (.05) & 8.21 (.03) & 8.26 (.10) & 8.38 (.03) & 8.64 (.04) & \textbf{2.36 (.37)} \\
 & 3.0 & 0.3 & 8.82 (.04) & 8.81 (.03) & 8.65 (.03) & 8.64 (.03) & 8.79 (.06) & 9.13 (.03) & \textbf{1.69 (.04)} \\
 & & 3.0 & 11.6 (.03) & 11.6 (.03) & 11.4 (.01) & 11.4 (.03) & 11.5 (.03) & 12.0 (.06) & \textbf{2.39 (.11)} \\
	\midrule
	\multicolumn{10}{c}{\textbf{d = 2}} \\
	\midrule
 0.3 & 0.3 & 0.3 & 2.39 (.03) & 2.41 (.04) & 2.06 (.01) & 2.13 (.02) & 2.90 (.03) & 2.94 (.06) & \textbf{1.22 (.01)} \\
 & & 3.0 & 4.49 (.02) & 4.48 (.03) & 4.21 (.02) & 4.17 (.03) & 4.95 (.04) & 5.03 (.06) & \textbf{2.08 (.05)} \\
 & 3.0 & 0.3 & 4.85 (.01) & 4.92 (.01) & 4.60 (.03) & 4.56 (.01) & 5.31 (.03) & 5.43 (.03) & \textbf{1.65 (.01)} \\
 & & 3.0 & 6.97 (.02) & 7.01 (.02) & 6.70 (.05) & 6.64 (.02) & 7.19 (.06) & 7.64 (.07) & \textbf{2.35 (.03)} \\
	\midrule
 3.0 & 0.3 & 0.3 & 5.66 (.02) & 5.70 (.03) & 5.38 (.03) & 5.41 (.03) & 5.63 (.06) & 6.26 (.06) & \textbf{1.30 (.01)} \\
 & & 3.0 & 8.41 (.03) & 8.46 (.02) & 8.09 (.02) & 8.08 (.03) & 8.15 (.04) & 9.18 (.05) & \textbf{2.10 (.01)} \\
 & 3.0 & 0.3 & 8.93 (.04) & 8.91 (.03) & 8.57 (.04) & 8.55 (.01) & 8.57 (.05) & 9.61 (.07) & \textbf{1.87 (.11)} \\
 & & 3.0 & 11.7 (.03) & 11.8 (.02) & 11.3 (.04) & 11.3 (.03) & 11.0 (.04) & 12.4 (.05) & \textbf{2.48 (.06)} \\
\bottomrule
\end{tabular}
  \end{adjustbox}
\end{table}

\begin{table}[H]
    \caption{Simulated datasets with longitudinal features: Mean running times (seconds) for $n = 100000$ observations, $p = 100$ fixed features, $q = 1000$ subjects, and $K = 3$ polynomial terms on $t$, random mode. All DNN experiments ran with 200 epochs.}
    \label{app:tab:additional:simulated:longitudinal:time}
    \begin{adjustbox}{width=\columnwidth,center}
    \centering
    \begin{tabular}{lll|lllllll}
\toprule
$\sigma^2_{b_0}$ & $\sigma^2_{b_1}$ & $\sigma^2_{b_2}$ & PCA-Ignore & PCA-OHE & VAE-Ignore & VAE-Embed & SVGPVAE & VRAE & LMMVAE \\
    \midrule
    \multicolumn{10}{c}{\textbf{d = 1}} \\
	\midrule
    0.3 & 0.3 & 0.3 & 0.8 & 10.6 & 81.1 & 87.2 & 407.4 & 676.0 & 191.4 \\
    &  & 3.0 & 0.7 & 10.4 & 80.8 & 87.1 & 405.9 & 673.2 & 191.5 \\
    & 3.0 & 0.3 & 0.8 & 10.3 & 80.7 & 86.9 & 406.6 & 677.5 & 191.4 \\
    &  & 3.0 & 0.8 & 10.4 & 81.2 & 87.2 & 405.7 & 673.5 & 192.0 \\
	\midrule
    3.0 & 0.3 & 0.3 & 0.7 & 10.3 & 79.9 & 86.6 & 405.6 & 652.0 & 190.4 \\
    &  & 3.0 & 0.9 & 10.5 & 80.0 & 86.6 & 407.6 & 661.9 & 191.1 \\
    & 3.0 & 0.3 & 0.7 & 10.3 & 80.5 & 86.7 & 410.6 & 658.1 & 191.8 \\
    &  & 3.0 & 0.7 & 10.6 & 80.4 & 86.5 & 409.2 & 658.7 & 191.6 \\
	\midrule
	\multicolumn{10}{c}{\textbf{d = 2}} \\
	\midrule
    0.3 & 0.3 & 0.3 & 0.8 & 10.7 & 83.5 & 89.2 & 725.5 & 671.0 & 194.3 \\
    &  & 3.0 & 0.8 & 11.0 & 83.1 & 89.3 & 726.7 & 671.1 & 193.7 \\
    & 3.0 & 0.3 & 0.9 & 10.4 & 83.0 & 88.8 & 726.4 & 672.0 & 193.9 \\
    &  & 3.0 & 0.8 & 10.5 & 83.1 & 89.2 & 736.2 & 665.9 & 194.6 \\
	\midrule
    3.0 & 0.3 & 0.3 & 0.8 & 10.5 & 82.1 & 88.1 & 728.0 & 658.9 & 193.0 \\
    &  & 3.0 & 0.8 & 10.6 & 83.0 & 88.6 & 731.0 & 660.4 & 194.2 \\
    & 3.0 & 0.3 & 0.7 & 10.4 & 82.9 & 88.8 & 733.4 & 655.8 & 194.3 \\
    &  & 3.0 & 0.8 & 10.5 & 83.1 & 88.7 & 733.3 & 660.6 & 194.5 \\
\bottomrule
\end{tabular}
    \end{adjustbox}
\end{table}

\begin{table}[H]
    \caption{Simulated datasets with longitudinal features: mean test NLL loss for $n = 100000$ observations, $p = 100$ fixed features, $q = 1000$ subjects, and $K = 3$ polynomial terms on $t$.}
    \label{app:tab:additional:simulated:longitudinal:nll}
    \begin{adjustbox}{width=\columnwidth,center}
    \centering
    \begin{tabular}{lll|lll|lll}
\toprule
 & & & \multicolumn{3}{c|}{\textbf{Random mode}} & \multicolumn{3}{c}{\textbf{Future mode}}\\
 \midrule
$\sigma^2_{b_0}$ & $\sigma^2_{b_1}$ & $\sigma^2_{b_2}$ & VAE-Ignore & VAE-Embed & LMMVAE & VAE-Ignore & VAE-Embed & LMMVAE \\
    \midrule
    \multicolumn{9}{c}{\textbf{d = 1}} \\
	\midrule
    0.3 & 0.3 & 0.3 & 160 & 160.6 & \textbf{106.1} & 209.3 & 207.2 & \textbf{99.3} \\
    &  & 3.0 & 229.2 & 226.8 & \textbf{119.1} & 417.0 & 424.2 & \textbf{110.8} \\
    & 3.0 & 0.3 & 268.7 & 269.6 & \textbf{123.5} & 457.3 & 462.8 & \textbf{118.6} \\
    &  & 3.0 & 338.4 & 336.7 & \textbf{135.6} & 665.6 & 666.3 & \textbf{153.0} \\
	\midrule
    3.0 & 0.3 & 0.3 & 459.7 & 460.2 & \textbf{115.9} & 538.8 & 536.8 & \textbf{110.0} \\
    &  & 3.0 & 553.7 & 553.7 & \textbf{116.5} & 820.8 & 826.2 & \textbf{160.7} \\
    & 3.0 & 0.3 & 611.5 & 613.4 & \textbf{114.8} & 865.6 & 864.8 & \textbf{121.1} \\
    &  & 3.0 & 708.0 & 704.1 & \textbf{128.9} & 1138.3 & 1143.0 & \textbf{156.8} \\
	\midrule
	\multicolumn{9}{c}{\textbf{d = 2}} \\
	\midrule
    0.3 & 0.3 & 0.3 & 162.9 & 165.5 & \textbf{106.8} & 206.3 & 213.3 & \textbf{99.4} \\
    &  & 3.0 & 230.0 & 230.1 & \textbf{126.9} & 421.2 & 417.6 & \textbf{117.8} \\
    & 3.0 & 0.3 & 270.6 & 272.2 & \textbf{120.3} & 460.6 & 456.4 & \textbf{118.4} \\
    &  & 3.0 & 335.5 & 340.4 & \textbf{135.2} & 670.1 & 664.7 & \textbf{161.6} \\
	\midrule
    3.0 & 0.3 & 0.3 & 462.0 & 465.1 & \textbf{123.4} & 538.1 & 540.9 & \textbf{107.8} \\
    &  & 3.0 & 557.7 & 552.7 & \textbf{145.6} & 809.3 & 807.9 & \textbf{127.7} \\
    & 3.0 & 0.3 & 606.6 & 609.1 & \textbf{125.0} & 857.7 & 855.0 & \textbf{138.8} \\
    &  & 3.0 & 691.7 & 691.5 & \textbf{135.6} & 1131.7 & 1126.2 & \textbf{161.4} \\
\bottomrule
\end{tabular}
    \end{adjustbox}
\end{table}

\begin{table}[H]
    \caption{Simulated datasets with spatial features: Mean reconstruction errors for $n = 100000$ observations, $p = 100$ fixed features, $q = 10000$ locations, random mode. Standard errors in parentheses, bold results are non-inferior to the best result in a paired t-test.}
    \label{app:tab:additional:simulated:spatial:random:mse}
    \centering
    \begin{tabular}{ll|llllll}
\toprule
$\sigma^2_{b}$ & $l^2$ & PCA-Ignore & PCA-OHE & VAE-Ignore & VAE-Embed. & SVGPVAE & LMMVAE \\
    \midrule
    \multicolumn{8}{c}{\textbf{d = 1}} \\
	\midrule
    0.3 & 0.3 & 1.49 (.02) & 1.47 (.02) & 1.31 (.05) & 1.31 (.01) & 1.58 (.05) & \textbf{1.02 (.00)} \\
    & 3.0 & 1.24 (.02) & 1.25 (.02) & 1.08 (.01) & 1.12 (.02) & 1.52 (.06) & \textbf{1.00 (.00)} \\
    3.0 & 0.3 & 3.53 (.11) & 3.53 (.04) & 1.65 (.02) & 1.73 (.02) & 2.45 (.07) & \textbf{1.20 (.01)} \\
    & 3.0 & 1.72 (.04) & 1.84 (.05) & 1.33 (.02) & 1.37 (.02) & 1.57 (.03) & \textbf{1.04 (.02)} \\
	\midrule
    \multicolumn{8}{c}{\textbf{d = 2}} \\
    \midrule
    0.3 & 0.3 & 1.71 (.05) & 1.66 (.03) & 1.28 (.04) & 1.41 (.04) & 1.92 (.02) & \textbf{1.10 (.03)} \\
    & 3.0 & 1.42 (.01) & 1.42 (.01) & 1.09 (.01) & 1.22 (.03) & 2.01 (.04) & \textbf{1.04 (.02)} \\
    3.0 & 0.3 & 3.37 (.07) & 3.30 (.03) & 1.66 (.05) & 1.68 (.03) & 2.44 (.03) & \textbf{1.22 (.02)} \\
    & 3.0 & 1.92 (.06) & 1.90 (.02) & 1.36 (.05) & 1.52 (.07) & 1.93 (.01) & \textbf{1.04 (.01)} \\
\bottomrule
\end{tabular}

\end{table}

\begin{table}[H]
    \caption{Simulated datasets with spatial features: Mean reconstruction errors for $n = 100000$ observations, $p = 100$ fixed features, $q = 10000$ locations, unknown mode. Standard errors in parentheses, bold results are non-inferior to the best result in a paired t-test.}
    \label{app:tab:additional:simulated:spatial:future:mse}
    \centering
    \begin{tabular}{ll|llllll}
\toprule
$\sigma^2_{b}$ & $l^2$ & PCA-Ignore & PCA-OHE & VAE-Ignore & VAE-Embed. & SVGPVAE & LMMVAE \\
    \midrule
    \multicolumn{8}{c}{\textbf{d = 1}} \\
	\midrule
    0.3 & 0.3 & 1.46 (.02) & 1.48 (.03) & 1.23 (.01) & 1.31 (.01) & 1.85 (.03) & \textbf{1.04 (.01)} \\
    & 3.0 & 1.25 (.03) & 1.24 (.02) & \textbf{1.12} (.05) & \textbf{1.10} (.01) & 1.56 (.03) & \textbf{1.09 (.08)} \\
    3.0 & 0.3 & 3.38 (.08) & 3.65 (.05) & 1.75 (.05) & 2.80 (.08) & 6.94 (.42) & \textbf{1.40 (.06)} \\
    & 3.0 & 1.78 (.05) & 1.79 (.02) & 1.41 (.04) & 1.52 (.03) & 2.58 (.09) & \textbf{1.14 (.02)} \\
    \midrule
    \multicolumn{8}{c}{\textbf{d = 2}} \\
    \midrule
    0.3 & 0.3 & 1.68 (.03) & 1.67 (.04) & 1.33 (.04) & 1.35 (.01) & 2.36 (.04) & \textbf{1.10 (.03)} \\
    & 3.0 & 1.43 (.01) & 1.43 (.02) & 1.11 (.03) & 1.15 (.02) & 1.96 (.06) & \textbf{1.04 (.03)} \\
    3.0 & 0.3 & 3.25 (.02) & 3.32 (.10) & 1.73 (.03) & 2.68 (.08) & 28.2 (5.8) & \textbf{1.37 (.04)} \\
    & 3.0 & 1.96 (.07) & 1.88 (.05) & 1.35 (.04) & 1.68 (.08) & 8.27 (3.2) & \textbf{1.12 (.03)} \\
\bottomrule
\end{tabular}

\end{table}

\begin{table}[H]
    \caption{Simulated datasets with spatial features: Mean running times (seconds) for $n = 100000$ observations, $p = 100$ fixed features, $q = 10000$ locations, random mode. All DNN experiments ran with 200 epochs.}
    \label{app:tab:additional:simulated:spatial:random:time}
    \centering
    \begin{tabular}{ll|llllll}
\toprule
$\sigma^2_{b}$ & $l^2$ & PCA-Ignore & PCA-OHE & VAE-Ignore & VAE-Embed. & SVGPVAE & LMMVAE \\
    \midrule
    \multicolumn{8}{c}{\textbf{d = 1}} \\
	\midrule
    0.3 & 0.3 & 1.7 & 38.4 & 94.5 & 112.8 & 495.3 & 192.1 \\
    & 3.0 & 1.2 & 36.9 & 94.1 & 111.7 & 497.6 & 190.1 \\
    3.0 & 0.3 & 1.1 & 36.9 & 93.8 & 112.4 & 497.1 & 190.9 \\
    & 3.0 & 1.2 & 36.8 & 93.8 & 111.7 & 494.4 & 190.1 \\
    \midrule
    \multicolumn{8}{c}{\textbf{d = 2}} \\
    \midrule
    0.3 & 0.3 & 1.2 & 37.4 & 96.7 & 114.3 & 949.2 & 192.8 \\
    & 3.0 & 1.3 & 37.1 & 96.0 & 113.8 & 939.6 & 193.8 \\
    3.0 & 0.3 & 1.4 & 37.2 & 96.3 & 113.8 & 933.9 & 191.2 \\
    & 3.0 & 1.6 & 37.6 & 96.1 & 114.0 & 935.6 & 190.8 \\
\bottomrule
\end{tabular}

\end{table}

\begin{table}[H]
    \caption{Simulated datasets with spatial features: mean test NLL loss for $n = 100000$ observations, $p = 100$ fixed features, $q = 10000$ locations, random mode.}
    \label{app:tab:additional:simulated:spatial:random:loss}
    \centering
    \begin{tabular}{ll|lll}
\toprule
$\sigma^2_{b}$ & $l^2$ & VAE-Ignore & VAE-Embed. & LMMVAE \\
    \midrule
    \multicolumn{5}{c}{\textbf{d = 1}} \\
	\midrule
    0.3 & 0.3 & \textbf{131.6} & \textbf{131.5} & \textbf{122.0} \\
    & 3.0 & 108.6 & 112.5 & \textbf{105.9} \\
    3.0 & 0.3 & \textbf{165.3} & \textbf{172.6} & \textbf{159.4} \\
    & 3.0 & \textbf{132.7} & 136.8 & 156.6 \\
    \midrule
    \multicolumn{5}{c}{\textbf{d = 2}} \\
    \midrule
    0.3 & 0.3 & \textbf{128.8} & 141.8 & \textbf{131.8} \\
    & 3.0 & \textbf{109.6} & 123.1 & \textbf{109.3} \\
    3.0 & 0.3 & \textbf{165.8} & \textbf{168.5} & \textbf{166.2} \\
    & 3.0 & \textbf{136.0} & 152.3 & 153.5 \\
\bottomrule
\end{tabular}

\end{table}

\newpage
\section{Real Data: Additional Results}
\label{app:additional:real}

\begin{table}[H]
    \caption{Real datasets with high-cardinality categorical features: mean test reconstruction errors over 5-CV runs. Standard errors in parentheses, bold results are non-inferior to the best result in a paired t-test.}
    \label{app:tab:additional:real:categorical}
    \centering
    \begin{tabular}{l|l|llll}
\toprule
Dataset & $d$ & PCA-Ignore & VAE-Ignore & VAE-Embed. & LMMVAE \\
    \midrule
    News & 1 & 0.97 (.00) & 0.81 (.00) & 0.85 (.00) & \textbf{0.75 (.00)} \\
    & 2 & 0.95 (.00) & 0.68 (.00) & 0.75 (.00) & \textbf{0.64 (.00)} \\
    & 5 & 0.91 (.00) & 0.54 (.00) & 0.63 (.00) & \textbf{0.51 (.00)} \\
    \midrule
    Spotify & 1 &  0.85 (.00) & 0.63 (.00) & 0.81 (.01) & \textbf{0.57 (.00)} \\
    & 2 & 0.74 (.00) & \textbf{0.49 (.01)} & 0.69 (.00) & \textbf{0.48 (.01)} \\
    & 5 & 0.49 (.00) & \textbf{0.20 (.00)} & 0.39 (.01) & \textbf{0.20 (.00)} \\
\bottomrule
\end{tabular}

\end{table}

\begin{table}[H]
    \caption{Real datasets with high-cardinality categorical features: mean running times (seconds) over 5-CV runs. All DNN experiments ran with 200 epochs.}
    \label{app:tab:additional:real:categorical:time}
    \centering
    \begin{tabular}{l|l|llll}
\toprule
Dataset & $d$ & PCA-Ignore & VAE-Ignore & VAE-Embed. & LMMVAE \\
    \midrule
    News & 1 & 1.7 & 91.2 & 127.9 & 610.5 \\
    & 2 & 1.7 & 93.4 & 129.6 & 619.0 \\
    & 5 & 1.8 & 94.1 & 130.8 & 613.4 \\
    \midrule
    Spotify & 1 & 0.1 & 32.9 & 85.6 & 86.8 \\
    & 2 & 0.1 & 33.7 & 82.7 & 88.8 \\
    & 5 & 0.1 & 34.3 & 83.3 & 88.1 \\
\bottomrule
\end{tabular}

\end{table}

\begin{table}[H]
    \caption{Real datasets with high-cardinality categorical features: mean test NLL loss over 5-CV runs.}
    \label{app:tab:additional:real:categorical:nll}
    \centering
    \begin{tabular}{l|l|llll}
\toprule
Dataset & $d$ & VAE-Ignore & VAE-Embed. & LMMVAE \\
    \midrule
    News & 1 & 143.1 & 150.0 & \textbf{39.7} \\
    & 2 & 120.5 & 133.0 & \textbf{35.8} \\
    & 5 & 95.1 & 111.0 & \textbf{31.7} \\
    \midrule
    Spotify & 1 & 9.5 & 12.2 & \textbf{2.9} \\
    & 2 & 7.5 & 10.5 & \textbf{2.5} \\
    & 5 & 3.2 & 6.1 & \textbf{1.4} \\
\bottomrule
\end{tabular}

\end{table}

\begin{table}[H]
    \caption{Real datasets with longitudinal features: mean test reconstruction errors over 5-CV runs. Standard errors in parentheses, bold results are non-inferior to the best result in a paired t-test.}
    \label{app:tab:additional:real:longitudinal}
    \begin{adjustbox}{width=\columnwidth,center}
    \centering
    \begin{tabular}{l|l|lllllll}
\toprule
Dataset & $d$ & PCA-Ignore & PCA-OHE & VAE-Ignore & VAE-Embed & SVGPVAE & VRAE & LMMVAE \\
    \midrule
    \multicolumn{8}{c}{\textbf{Random mode}} \\
    \midrule
    Rossmann & 1 & 0.89 (.01) & 0.89 (.01) & 0.44 (.01) & 0.43 (.01) &  0.57 (.00) & 1.28 (.01) & \textbf{0.04 (.00)} \\
    & 2 & 0.78 (.01) & 0.79 (.01) & 0.20 (.01) & 0.21 (.01) &  0.45 (.00) & 1.28 (.01) & \textbf{0.01 (.00)} \\
    & 5 & 0.54 (.00) & 0.54 (.00) & 0.03 (.00) & 0.03 (.01) &  0.39 (.00) & 1.28 (.01) & \textbf{0.005 (.00)} \\
    \midrule
    UKB & 1 & 0.93 (.001) & -- & 0.77 (.006) & 0.84 (.010) & -- & -- & \textbf{0.76 (.005)} \\
    & 2 & 0.87 (.001) & -- & 0.64 (.003) & 0.75 (.005) & -- & -- & \textbf{0.63 (.002)} \\
    & 5 & 0.74 (.001) & -- & 0.38 (.004) & 0.53 (.009) & -- & -- & \textbf{0.37 (.002)} \\
    \midrule
    \multicolumn{8}{c}{\textbf{Future mode}} \\
    \midrule
    Rossmann & 1 & 0.92 (.00) & 0.91 (.00) & 0.54 (0.02) & 0.56 (.02) & 0.63 (.01) & 1.16 (.00) & \textbf{0.06 (.00)} \\
    & 2 & 0.79 (.00) & 0.80 (.00) & 0.29 (.01) & 0.30 (.03) & 0.49 (.00) & 1.16 (.01) & \textbf{0.05 (.01)} \\
    & 5 & 0.60 (.00) & 0.61 (.00) & 0.05 (.00) & 0.05 (.00) & 0.97 (.09) & 1.17 (.01) & \textbf{0.03 (.00)} \\
    \midrule
    UKB & 1 & 0.94 (.000) & -- & 0.81 (.015) & 0.86 (.007) & -- & -- & \textbf{0.78 (.004)} \\
    & 2 & 0.88 (.000) & -- & 0.67 (.005) & 0.79 (.006) & -- & -- & \textbf{0.66 (.004)} \\
    & 5 & 0.75 (.000) & -- & \textbf{0.42 (.002)} & 0.59 (.014) & -- & -- & \textbf{0.41 (.006)} \\
\bottomrule
\end{tabular}

    \end{adjustbox}
\end{table}

\begin{table}[H]
    \caption{Real datasets with longitudinal features: mean running times (seconds) over 5-CV runs. All DNN experiments ran with 200 epochs.}
    \label{app:tab:additional:real:longitudinal:time}
    \begin{adjustbox}{width=\columnwidth,center}
    \centering
    \begin{tabular}{l|l|lllllll}
\toprule
Dataset & $d$ & PCA-Ignore & PCA-OHE & VAE-Ignore & VAE-Embed & SVGPVAE & VRAE & LMMVAE \\
    \midrule
    \multicolumn{8}{c}{\textbf{Random mode}} \\
    \midrule
    Rossmann & 1 & 0.1 & 1.5 & 36.7 & 57.7 & 152.7 & 61.2 & 72.4 \\
    & 2 & 0.1 & 1.5 & 37.4 & 50.9 & 264.4 & 59.5 & 66.3 \\
    & 5 & 0.1 & 1.6 & 40.1 & 51.4 & 283.7 & 59.5 & 66.7 \\
    \midrule
    UKB & 1 & 1.6 & -- & 443.4 & 592.5 & -- & -- & 8149 \\
    & 2 & 1.7 & -- & 459.1 & 601.0 & -- & -- & 8122 \\
    & 5 & 1.5 & -- & 457.6 & 603.3 & -- & -- & 8256 \\
    \midrule
    \multicolumn{8}{c}{\textbf{Future mode}} \\
    \midrule
    Rossmann & 1 & 0.1 & 1.3 & 25.0 & 27.5 & 122.8 & 59.4 & 45.7 \\
    & 2 & 0.1 & 1.2 & 25.8 & 28.2 & 213.0 & 59.1 & 45.8 \\
    & 5 & 0.1 & 1.3 & 25.9 & 28.2 & 483.1 & 59.3 & 45.7 \\
    \midrule
    UKB & 1 & 1.2 & -- & 354.1 & 474.4 & -- & -- & 6585 \\
    & 2 & 1.4 & -- & 364.0 & 482.7 & -- & -- & 6601 \\
    & 5 & 1.1 & -- & 365.4 & 484.7 & -- & -- & 6608 \\
\bottomrule
\end{tabular}

    \end{adjustbox}
\end{table}

\begin{table}[H]
    \caption{Real datasets with longitudinal features: mean test NLL loss over 5-CV runs.}
    \label{app:tab:additional:real:longitudinal:nll}
    \centering
    \begin{tabular}{l|l|lll}
\toprule
Dataset & $d$ & VAE-Ignore & VAE-Embed & LMMVAE \\
    \midrule
    \multicolumn{5}{c}{\textbf{Random mode}} \\
    \midrule
    Rossmann & 1 & 9.8 & 9.8 & \textbf{1.4}\\
    & 2 & 4.7 & 4.9 & \textbf{1.3} \\
    & 5 & \textbf{0.8} & \textbf{1.0} & \textbf{1.3} \\
    \midrule
    UKB & 1 & 41.1 & 42.1 & \textbf{31.7}\\
    & 2 & 31.7 & 37.2 & \textbf{26.6} \\
    & 5 & 19.3 & 27.4 & \textbf{16.4} \\
    \midrule
    \multicolumn{5}{c}{\textbf{Future mode}} \\
    \midrule
    Rossmann & 1 & 12.3 & 12.3 & \textbf{2.6}\\
    & 2 & 6.8 & 6.8 & \textbf{2.1} \\
    & 5 & \textbf{1.6} & \textbf{1.6} & \textbf{2.0} \\
    \midrule
    UKB & 1 & 38.9 & 46.9 & \textbf{14.6}\\
    & 2 & 32.6 & 40.3 & \textbf{11.8} \\
    & 5 & 21.1 & 31.0 & \textbf{7.8} \\
\bottomrule
\end{tabular}

\end{table}

\begin{table}[H]
    \caption{Real datasets with spatial features: mean test reconstruction errors over 5-CV runs. Standard errors in parentheses, bold results are non-inferior to the best result in a paired t-test.}
    \label{app:tab:additional:real:spatial}
    \begin{adjustbox}{width=\columnwidth,center}
    \centering
    \begin{tabular}{l|l|lllllll}
\toprule
Dataset & $d$ & PCA-Ignore & PCA-OHE & VAE-Ignore & VAE-OHE & VAE-Embed. & SVGPVAE & LMMVAE \\
    \midrule
    \multicolumn{8}{c}{\textbf{Random mode}} \\
    \midrule
    Income & 1 & .016 (.000) &.016 (.000) & .018 (.000) & .019 (.000) & .018 (.000) & .018 (.000) & \textbf{.008 (.000)} \\
    & 2 & .014 (.000) & .014 (.000) & .012 (.000) & .019 (.000) & .011 (.000) & .017 (.000) & \textbf{.004 (.000)} \\
    & 5 & .010 (.000) & .010 (.000) & \textbf{.003 (.000)} & .019 (.000) & \textbf{.003 (.000)} & .016 (.000) & \textbf{.003 (.000)} \\
    \midrule
    Asthma & 1 & .017 (.000) & .017 (.000) & .025 (.000) & .025 (.000) & .024 (.000) & .024 (.000) & \textbf{.012 (.000)} \\
    & 2 & .015 (.000) & .015 (.000) & .016 (.000) & .025 (.000) & .016 (.001) & .022 (.000) & \textbf{.006 (.000)} \\
    & 5 & .011 (.000) & .011 (.000) & \textbf{.003 (.000)} & .025 (.000) & \textbf{.003 (.000)} & .026 (.002) & \textbf{.005 (.000)} \\
    \midrule
    \multicolumn{8}{c}{\textbf{Unknown mode}} \\
    \midrule
    Income & 1 & .016 (.000) & .016 (.000) & .019 (.000) & .019 (.000) & .021 (.001) & .022 (.001) & \textbf{.008 (.000)} \\
    & 2 & .014 (.000) & .014 (.000) & .015 (.001) & .019 (.000) &  .027 (.000) & .022 (.001) & \textbf{.005 (.001)} \\
    & 5 & .010 (.000) & .010 (.000) & .003 (.000) & .019 (.000) &  .006 (.000) & .037 (.001) & \textbf{.003 (.000)} \\
    \midrule
    Asthma & 1 & .017 (.001) & .017 (.001) & .026 (.001) & .026 (.001) & .027 (.001) & .028 (.001) & \textbf{.012 (.001)} \\
    & 2 & .016 (.000) & .016 (.000) & .020 (.001) & .025 (.001) & .035 (.003) & .030 (.001) & \textbf{.006 (.000)} \\
    & 5 & .011 (.000) & .011 (.000) & \textbf{.004 (.001)} & .025 (.001) & .006 (.000) & .059 (.003) & \textbf{.004 (.000)} \\
\bottomrule
\end{tabular}

    \end{adjustbox}
\end{table}

\begin{table}[H]
    \caption{Real datasets with spatial features: mean running times (seconds) over 5-CV runs. All DNN experiments ran with 200 epochs.}
    \label{app:tab:additional:real:spatial:time}
    \begin{adjustbox}{width=\columnwidth,center}
    \centering
    \begin{tabular}{l|l|lllllll}
\toprule
Dataset & $d$ & PCA-Ignore & PCA-OHE & VAE-Ignore & VAE-OHE & VAE-Embed. & SVGPVAE & LMMVAE \\
    \midrule
    \multicolumn{8}{c}{\textbf{Random mode}} \\
    \midrule
    Income & 1 &  0.2 & 8.1 & 59.3 & 256.4 & 65.0 & 304.2 & 112.1 \\
    & 2 & 0.2 & 8.1 & 61.5 & 259.9 & 66.9 & 536.8 & 111.3 \\
    & 5 & 0.2 & 8.1 & 61.9 & 263.0 & 67.4 & 1234.9 & 111.9 \\
    \midrule
    Asthma & 1 & 0.2 & 7.9 & 57.1 & 245.9 & 62.3 & 294.9 & 106.8 \\
    & 2 & 0.2 & 8.1 & 58.6 & 246.6 & 63.6 & 517.8 & 106.7 \\
    & 5 & 0.1 & 7.7 & 58.9 & 246.1 & 63.7 & 1194.8 & 107.5 \\
    \midrule
    \multicolumn{8}{c}{\textbf{Unknown mode}} \\
    \midrule
    Income & 1 & 0.2 & 7.2 & 58.8 & 219.2 & 64.2 & 300.0 & 107.5 \\
    & 2 & 0.2 & 7.3 & 60.9 & 222.2 & 66.1 & 532.4 & 110.3\\
    & 5 & 0.1 & 7.3 & 60.8 & 225.2 & 66.2 & 1249.2 & 110.0 \\
    \midrule
    Asthma & 1 & 0.2 & 6.6 & 56.9 & 211.2 & 62.1 & 296.5 & 105.5 \\
    & 2 & 0.2 & 6.5 & 58.5 & 213.7 & 63.5 & 518.6 & 108.4 \\
    & 5 & 0.2 & 6.4 & 59.2 & 215.1 & 64.1 & 1190.5 & 106.3 \\
\bottomrule
\end{tabular}

    \end{adjustbox}
\end{table}

\begin{table}[H]
    \caption{Real datasets with spatial features: mean test NLL over 5-CV runs.}
    \label{app:tab:additional:real:spatial:nll}
    \centering
    \begin{tabular}{l|l|lll}
\toprule
Dataset & $d$ & VAE-Ignore & VAE-Embed. & LMMVAE \\
    \midrule
    \multicolumn{5}{c}{\textbf{Random mode}} \\
    \midrule
    Income & 1 & 2.2 & 3.1 & \textbf{0.3} \\
    & 2 & 0.5 & 0.5 & \textbf{0.2} \\
    & 5 & 0.2 & 0.2 & \textbf{0.1} \\
    \midrule
    Asthma & 1 & 2.5 & 4.9 & \text{0.4} \\
    & 2 & 0.7 & 0.9 & \textbf{0.2} \\
    & 5 & 0.3 & 0.3 & \textbf{0.2} \\
    \midrule
    \multicolumn{5}{c}{\textbf{Unknown mode}} \\
    \midrule
    Income & 1 & 3.0 & 13.6 & \textbf{0.3} \\
    & 2 & 0.6 & 1.1 & \textbf{0.2} \\
    & 5 & 0.2 & 0.4 & \textbf{0.1} \\
    \midrule
    Asthma & 1 & 2.8 & 14.6 & \textbf{0.4} \\
    & 2 & 0.8 & 2.3 & \textbf{0.2} \\
    & 5 & 0.3 & 0.4 & \textbf{0.2} \\
\bottomrule
\end{tabular}

\end{table}

\begin{table}[H]
    \caption{Real datasets with spatial and categorical features: mean test reconstruction errors over 5-CV runs. Standard errors in parentheses, bold results are non-inferior to the best result in a paired t-test.}
    \label{app:tab:additional:real:spatial-categorical:mse}
    \centering
    \begin{tabular}{l|l|llll}
\toprule
Dataset & $d$ & PCA-Ignore & VAE-Ignore & VAE-Embed. & LMMVAE \\
    \midrule
    \multicolumn{6}{c}{\textbf{Random mode}} \\
    \midrule
    Cars & 1 & .087 (.003) & .086 (.001) & .090 (.000) & \textbf{.036 (.000)} \\
    & 2 & .081 (.003) & .065 (.000) & .068 (.002) & \textbf{.028 (.001)} \\
    & 5 & .069 (.004) & .026 (.000) & .028 (.000) & \textbf{.008 (.001)} \\
    \midrule
    Airbnb & 1 & .066 (.000) & .060 (.000) & .061 (.000) & \textbf{.057 (.000)}  \\
    & 2 & .061 (.000) & .053 (.000) & .052 (.000) & \textbf{.050 (.000)} \\
    & 5 & .053 (.000) & .042 (.000) & .042 (.000) & \textbf{.039 (.000)} \\
    \midrule
    \multicolumn{6}{c}{\textbf{Unknown mode}} \\
    \midrule
    Cars & 1 & .087 (.003) & .088 (.001) & .109 (.005) & \textbf{.037 (.001)} \\
    & 2 & .081 (.003) & .071 (.001) & .125 (.003) & \textbf{.026 (.000)} \\
    & 5 & .069 (.003) & .026 (.001) & .030 (.001) & \textbf{.008 (.001)} \\
    \midrule
    Airbnb & 1 & .066 (.000) & .060 (.000) & .078 (.002) & \textbf{.056 (.000)} \\
    & 2 & .061 (.000) & .053 (.000) & .064 (.001) & \textbf{.050 (.000)} \\
    & 5 & .053 (.000) & .041 (.000) & .043 (.000) & \textbf{.040 (.000)} \\
\bottomrule
\end{tabular}

\end{table}

\begin{table}[H]
    \caption{Real datasets with spatial and categorical features: mean running times (seconds) over 5-CV runs. All DNN experiments ran with 200 epochs.}
    \label{app:tab:additional:real:spatial-categorical:time}
    \centering
    \begin{tabular}{l|l|llll}
\toprule
Dataset & $d$ & PCA-Ignore & VAE-Ignore & VAE-Embed. & LMMVAE \\
    \midrule
    \multicolumn{6}{c}{\textbf{Random mode}} \\
    \midrule
    Cars & 1 & 0.4 & 84.4 & 94.1 & 236.8 \\
    & 2 & 0.4 & 86.3 & 95.9 & 238.0 \\
    & 5 & 0.4 & 86.1 & 96.4 & 238.5 \\
    \midrule
    Airbnb & 1 & 0.4 & 51.2 & 55.9 & 238.4 \\
    & 2 & 0.5 & 52.6 & 56.8 & 238.6 \\
    & 5 & 0.5 & 51.8 & 57.2 & 238.7 \\
    \midrule
    \multicolumn{6}{c}{\textbf{Unknown mode}} \\
    \midrule
    Cars & 1 & 0.4 & 82.3 & 93.8 & 236.5 \\
    & 2 & 0.4 & 83.9 & 95.4 & 236.5 \\
    & 5 & 0.4 & 84.7 & 96.1 & 237.3 \\
    \midrule
    Airbnb & 1 & 0.4 & 48.8 & 55.5 & 250.3 \\
    & 2 & 0.4 & 49.7 & 56.5 & 248.7 \\
    & 5 & 0.4 & 50.1 & 56.8 & 247.5 \\
\bottomrule
\end{tabular}

\end{table}

\begin{table}[H]
    \caption{Real datasets with spatial and categorical features: mean test NLL loss over 5-CV runs.}
    \label{app:tab:additional:real:spatial-categorical:nll}
    \centering
    \begin{tabular}{l|l|lll}
\toprule
Dataset & $d$ & VAE-Ignore & VAE-Embed. & LMMVAE \\
    \midrule
    \multicolumn{5}{c}{\textbf{Random mode}} \\
    \midrule
    Cars & 1 & 7.2 & 7.7 & \textbf{2.0} \\
    & 2 & 5.2 & 5.4 & \textbf{1.9} \\
    & 5 & 2.1 & 2.3 & \textbf{1.0} \\
    \midrule
    Airbnb & 1 & 13.4 & 13.4 & \textbf{11.1} \\
    & 2 & 11.3 & 11.0 & \textbf{10.0} \\
    & 5 & 8.3 & 8.5 & \textbf{7.9} \\
    \midrule
    \multicolumn{5}{c}{\textbf{Unknown mode}} \\
    \midrule
    Cars & 1 & 7.8 & 13.3 & \textbf{2.0} \\
    & 2 & 5.7 & 10.1 & \textbf{1.6} \\
    & 5 & 2.2 & 2.5 & \textbf{1.0} \\
    \midrule
    Airbnb & 1 & 13.2 & 20.2 & \textbf{11.1} \\
    & 2 & 11.3 & 14.2 & \textbf{9.9} \\
    & 5 & 8.3 & 8.7 & \textbf{7.9} \\
\bottomrule
\end{tabular}

\end{table}

\begin{table}[H]
    \caption{The CelebA image dataset with a single high-cardinality categorical feature: mean test reconstruction error over 5-CV runs.}
    \label{app:tab:additional:real:image:mse}
    \centering
    \begin{tabular}{l|lll}
\toprule
$d$ & VAE-Ignore & VAE-Embed. & LMMVAE \\
    \midrule
    100 & .0512 (.008) & \textbf{.0091 (.000)} & \textbf{.0090 (.000)} \\
    200 & .0349 (.001) & \textbf{.0065 (.000)} & \textbf{.0064 (.000)} \\
    500 & .0354 (.001) & \textbf{.0043 (.000)} & \textbf{.0042 (.000)} \\
\bottomrule
\end{tabular}

\end{table}

\begin{table}[H]
    \caption{The CelebA image dataset with a single high-cardinality categorical feature: mean test NLL loss and running times (hours) over 5-CV runs.}
    \label{app:tab:additional:real:image:time_nll}
    \centering
    \begin{tabular}{l|lll|lll}
\toprule
& \multicolumn{3}{c|}{\textbf{NLL loss}} & \multicolumn{3}{c}{\textbf{Runtime (hours)}}\\
    \midrule
$d$ & VAE-Ignore & VAE-Embed. & LMMVAE & VAE-Ignore & VAE-Embed. & LMMVAE\\
    \midrule
    100 & 661.0 & \textbf{119.3} & \textbf{118.1} & 2.3 & 2.4 & 3.1 \\
    200 & 453.2 & \textbf{93.3} & \textbf{95.5} & 2.3 & 2.3 & 3.0 \\
    500 & 457.3 & \textbf{73.5} & \textbf{74.1} & 2.4 & 2.4 & 3.2 \\
\bottomrule
\end{tabular}

\end{table}

\end{document}